\documentclass{vldb}

\newcommand{\paperTitle}{ODIN: Automated Drift Detection and Recovery in Video Analytics}
\newcommand{\paperKeywords}{}
\newcommand{\paperAuthors}{Abhijit Suprem, Joy Arulraj, Calton Pu, Joao Ferreira}

\usepackage[hyphens]{url}
\usepackage{xstring}

\usepackage{etoolbox}
\usepackage{pifont}
\usepackage{amssymb}
\usepackage[hyphens]{url}
\usepackage[breaklinks,colorlinks]{hyperref}
\usepackage[usenames,dvipsnames]{xcolor}
\hypersetup{citecolor=blue,linkcolor=blue}
\usepackage{amsmath,amsopn,amssymb}
\usepackage{subcaption}
\usepackage{endnotes,microtype,xspace,graphicx,fancyvrb,multirow}
\usepackage{booktabs}
\usepackage{array,underscore,relsize}
\usepackage[T1]{fontenc}
\usepackage{times}
\usepackage{comment}
\usepackage{enumitem}
\usepackage{nicefrac}
\usepackage[labelfont=bf,font=small,skip=5pt]{caption}

\usepackage{fp}
\usepackage{siunitx}

\usepackage{balance}
\usepackage{nopageno}

\usepackage{floatrow}
\newfloatcommand{capbtabbox}{table}[][\FBwidth]

\usepackage[usenames,dvipsnames]{xcolor}
\definecolor{linkcolor}{HTML}{647382} 
\definecolor{citecolor}{HTML}{647382}           
\definecolor{urlcolor}{rgb}{0.4,0.2,0.2}
\definecolor{sqlcolor}{HTML}{965d67}
\definecolor{smtcolor}{HTML}{5d968c}

\usepackage{afterpage}
\usepackage{graphicx, balance, epsfig, endnotes}
\usepackage{colortbl}
\usepackage{tabularx}
\usepackage{listings}

\usepackage[breaklinks]{hyperref}
\hypersetup{%
    pdfauthor = {\paperAuthors},
    pdftitle = {\paperTitle},
    pdfkeywords = {\paperKeywords},
    bookmarksopen = {true},
    colorlinks=true,
    citecolor={urlcolor},
    linkcolor={linkcolor},
    urlcolor={citecolor},
    pdfborder={ 0 0 0 }
}

\usepackage{mathtools}
\usepackage{arydshln}

\usepackage[ruled,vlined,linesnumbered]{algorithm2e}

\usepackage[capitalize,noabbrev,nameinlink]{cleveref}

\crefname{algocf}{alg.}{algs.}
\Crefname{algocf}{Algorithm}{Algorithms}

\fvset{fontsize=\scriptsize,xleftmargin=8pt,numbers=left,numbersep=5pt}

\input{code/fmt}

\newcommand{\zerodisplayskips}{%
  \setlength{\abovedisplayskip}{0pt}%
  \setlength{\belowdisplayskip}{0pt}%
  \setlength{\abovedisplayshortskip}{0pt}%
  \setlength{\belowdisplayshortskip}{0pt}}
\appto{\normalsize}{\zerodisplayskips}
\appto{\small}{\zerodisplayskips}
\appto{\footnotesize}{\zerodisplayskips}

\def\Snospace~{\S{}}

\newcommand{\dcircle}[1]{\ding{\numexpr181 + #1}}

\newcommand{\x}{$\times$\xspace}

\newif\ifdraft\drafttrue
\newif\ifnotes\notestrue
\ifdraft\else\notesfalse\fi

\input{glyphtounicode}
\newcolumntype{R}[1]{>{\raggedleft\let\newline\\\arraybackslash\hspace{0pt}}p{#1}}

\newcommand{\includepdf}[1]{
  \includegraphics[width=\columnwidth]{#1}
}

\usepackage{siunitx}
\usepackage{hhline}

\usepackage{tikz}

\newcommand*\circled[2][1.6]{\tikz[baseline=(char.base)]{
    \node[shape=circle, draw, inner sep=1pt, 
        minimum height={\f@size*#1},] (char) {\vphantom{WAH1g}#2};}}
\makeatother

\SetCommentSty{mycommfont}

\usepackage{xstring}
\newcommand{\PP}[1]{
\vspace{2px}
\noindent{\bf \IfEndWith{#1}{.}{#1}{#1.}}
}

\usepackage{xstring}
\newcommand{\PPP}[1]{
\vspace{2px}
\indent{\it \IfEndWith{#1}{.}{#1}{#1.}}
}

\newcommand{\bdd}{BDD\xspace}
\newcommand{\mnist}{MNIST\xspace}
\newcommand{\cifar}{CIFAR-10\xspace}

\newcommand{\sys}{\textsc{Odin}\xspace}
\newcommand{\sysheavy}{\textsc{Odin-Heavy}\xspace}
\newcommand{\sysfilter}{\textsc{Odin-Filter}\xspace}
\newcommand{\syspp}{\textsc{Odin-PP}\xspace}
\newcommand{\detector}{\textsc{Detector}\xspace}
\newcommand{\specializer}{\textsc{Specializer}\xspace}
\newcommand{\selector}{\textsc{Selector}\xspace}
\newcommand{\manager}{\textsc{ModelManager}\xspace}

\newcommand{\yololocal}{\textsc{YOLO-specialized}\xspace}
\newcommand{\yolo}{\textsc{YOLO}\xspace}
\newcommand{\yololite}{\textsc{YOLO-lite}\xspace}

\newcommand{\cday}{\textsc{C-$\alpha$}\xspace}
\newcommand{\cnight}{\textsc{C-$\beta$}\xspace}
\newcommand{\crainy}{\textsc{C-$\gamma$}\xspace}
\newcommand{\csnowy}{\textsc{C-$\delta$}\xspace}

\newcommand{\cmday}{\textsc{C-$\alpha$}\xspace}
\newcommand{\cmnight}{\textsc{C-$\beta$}\xspace}
\newcommand{\cmrainy}{\textsc{C-$\gamma$}\xspace}
\newcommand{\cmsnowy}{\textsc{C-$\delta$}\xspace}

\newcommand{\fulldata}{\textsc{Full-Data}\xspace}
\newcommand{\daydata}{\textsc{Day-Data}\xspace}
\newcommand{\nightdata}{\textsc{Night-Data}\xspace}
\newcommand{\rainydata}{\textsc{Rain-Data}\xspace}
\newcommand{\snowydata}{\textsc{Snow-Data}\xspace}

\newcommand{\ie}{\textit{i}.\textit{e}.\xspace}
\newcommand{\eg}{\textit{e}.\textit{g}.\xspace}

\newcommand{\boxbeg}{
\vspace{5px}
\noindent\begin{tabular}{|l|}\hline
\begin{minipage}{3.2in}
\vspace{5px}
\noindent
}

\newcommand{\boxend}{
\vspace{5px}
\end{minipage}\\ \hline
\end{tabular}
\vspace{1px}
}

\newcommand{\ttt}{\texttt}

\makeatletter
\newcommand\BeraMonottfamily{%
  \def\fvm@Scale{0.85}
  \fontfamily{fvm}\selectfont
}
\makeatother
\lstdefinestyle{SQLStyle}{
language=SQL,
basicstyle=\BeraMonottfamily\scriptsize, 
keywordstyle=\color{sqlcolor}\bfseries,
literate = {-}{-}1, 
backgroundcolor = \color{lightgray!50},
aboveskip=2pt,belowskip=2pt
}
\lstdefinestyle{ScriStyle}{
    language=SQL,
    basicstyle=\BeraMonottfamily\footnotesize, 
    keywordstyle=\color{smtcolor}\bfseries,
    morekeywords={and, or, not},
    literate = {-}{-}1, 
}
\crefname{lstlisting}{listing}{listings}
\Crefname{lstlisting}{Listing}{Listings}

\definecolor{webgreen}{rgb}{0,.5,0}
\definecolor{webbrown}{rgb}{.6,0,0}
\definecolor{webblue}{rgb}{0,0,.7}

\newcommand{\squishitemize}{
 \begin{list}{$\bullet$}
  { \setlength{\itemsep}{0pt}
     \setlength{\parsep}{3pt}
     \setlength{\topsep}{3pt}
     \setlength{\partopsep}{0pt}
     \setlength{\leftmargin}{1.95em}
     \setlength{\labelwidth}{1.5em}
     \setlength{\labelsep}{0.5em} } }

\newcounter{Lcount}
\newcommand{\squishlist}{
    \begin{list}{\arabic{Lcount}. }
   { \usecounter{Lcount}
        \setlength{\itemsep}{0pt}
        \setlength{\parsep}{3pt}
        \setlength{\topsep}{3pt}
        \setlength{\partopsep}{0pt}
        \setlength{\leftmargin}{2em}
        \setlength{\labelwidth}{1.5em}
        \setlength{\labelsep}{0.5em} } }

\newcommand{\squishend}{\end{list}}

\newcommand{\edit}[1]{\textcolor{black}{#1}}

\usepackage[usenames,dvipsnames]{xcolor}
\definecolor{linkcolor}{HTML}{647382}
\definecolor{citecolor}{HTML}{647382} %
\definecolor{urlcolor}{rgb}{0.4,0.2,0.2}
\definecolor{sqlcolor}{HTML}{965d67}
\definecolor{smtcolor}{HTML}{5d968c}
\definecolor{webblue}{rgb}{0,0,.7}
\definecolor{webgreen}{rgb}{0,.5,0}
\definecolor{webbrown}{rgb}{.6,0,0}
\definecolor{Red}{rgb}{1,0,0}

\usepackage[breaklinks]{hyperref}
\hypersetup{%
	pdfauthor = {\paperAuthors},
	pdftitle = {\paperTitle},
	pdfkeywords = {\paperKeywords},
	bookmarksopen = {true},
	colorlinks=true,
	citecolor={urlcolor},
	linkcolor={linkcolor},
	urlcolor={citecolor},
	pdfborder={ 0 0 0 }
}

\usepackage[usenames,dvipsnames]{xcolor}
\usepackage{amsmath,amsopn,amssymb}
\usepackage{listings}
\usepackage{subcaption}
\usepackage{endnotes,microtype,xspace,graphicx,fancyvrb,multirow}
\usepackage{supertabular,booktabs}
\usepackage{array,underscore,relsize}
\usepackage[T1]{fontenc}
\usepackage{times}
\usepackage{fancyhdr,lastpage}
\usepackage{enumitem}
\usepackage{balance}
\usepackage{booktabs}
\usepackage{pifont}
\usepackage{listings}
\usepackage{multirow}
\usepackage[normalem]{ulem}
\useunder{\uline}{\ul}{}
\usepackage[scaled]{beramono}
\usepackage{tabularx}
\usepackage{wrapfig}
\usepackage{lipsum}

\usepackage{stmaryrd}
\usepackage{ltablex}
\usepackage{mathtools}
\usepackage{mathrsfs}
\usepackage{ dsfont }
\usepackage{pifont}

\usepackage[ruled,vlined,linesnumbered]{algorithm2e}

\vldbTitle{\paperTitle}
\vldbAuthors{\paperAuthors}
\vldbDOI{https://doi.org/10.14778/3407790.3407837}
\vldbVolume{13}
\vldbNumber{11}
\vldbYear{2020}

\SetAlFnt{\small}
\SetAlCapFnt{\small}
\SetAlCapNameFnt{\small}

\begin{document}


\newcommand{\mail}[1]{\href{mailto:#1}{#1}}

\author{
\def\arraystretch{1}
\begin{tabular}{cccc} 
 Abhijit Suprem$^1$ & Joy Arulraj$^1$ & Calton Pu$^1$ & Joao Ferreira$^2$ \\
 \mail{asuprem@gatech.edu} & \mail{arulraj@gatech.edu} &
 \mail{calton.pu@cc.gatech.edu} & \mail{jef@ime.usp.br}\\
 \multicolumn{4}{c}{$^1$\affaddr{Georgia Institute of Technology}, $^2$\affaddr{University of Sao Paulo}}\\
\end{tabular}
}

\title{\paperTitle}
\date{}

\maketitle

\begin{abstract}
Recent advances in computer vision have led to a resurgence of interest in
visual data analytics.
Researchers are developing systems for effectively and efficiently analyzing
visual data at scale.
A significant challenge that these systems encounter lies in the drift in real-world
visual data.
For instance, a model for self-driving vehicles that is not trained on images
containing snow does not work well when it encounters them in practice.
This drift phenomenon limits the accuracy of models employed for visual data
analytics.

In this paper, we present a visual data analytics system, called \sys, that
automatically detects and recovers from drift.
\sys uses adversarial autoencoders to learn the distribution of
high-dimensional images.
We present an unsupervised algorithm for detecting drift by comparing the
distributions of the given data against that of previously seen data.
When \sys detects drift, it invokes a drift recovery algorithm
to deploy specialized models tailored towards the novel data points.
These specialized models outperform their non-specialized counterpart on
accuracy, performance, and memory footprint.
Lastly, we present a model selection algorithm for picking an ensemble of
best-fit specialized models to process a given input.
We evaluate the efficacy and efficiency of \sys on high-resolution dashboard
camera videos captured under diverse environments from the Berkeley DeepDrive
dataset.
We demonstrate that \sys{'s} models deliver 6$\times$ higher throughput, 2$\times$ higher
accuracy, and 6$\times$ smaller memory footprint compared to a baseline system
without automated drift detection and recovery.
\end{abstract}

\section{Introduction} 
\label{sec:intro}

Recent advances in computer vision (\eg, image
classification~\cite{classsurvey}, object detection~\cite{detectsurvey}, and
object tracking~\cite{tracksurvey}) have led to a resurgence of interest in
visual data analytics.
Researchers are developing database management systems (DBMSs) for analyzing
visual data at scale~\cite{blazeit,videoedge,deeplens,chameleon}.
While these systems deliver high performance, they suffer from a major
limitation that constrains their accuracy on real-world visual data.
They assume that all the frames of videos stem from a static distribution.
In practice, the visual data \textit{drifts} over time because it comes from
a dynamic, time-evolving distribution.
For instance, a machine learning (ML) model for self-driving vehicles that is
not trained on images containing snow does not work well when it encounters them
in practice~\cite{bdd}.
This phenomenon is referred to as \textit{concept
drift}~\cite{drift2,drift3}, and it limits the efficacy of ML models employed
in visual DBMSs.

\PP{Challenges}
Concept drift is well studied in the domain of 
low-dimensional, structured data analysis~\cite{drift2}.
For instance, Kalman filtering is a widely-used technique for recovering from
data drift due to sensor failures~\cite{kalman}.
However, these techniques cannot cope with drift in high-dimensional,
unstructured data (\eg, images~\cite{dynaclassifier}, videos~\cite{videodrift}).
State-of-the-art ML models assume that the data comes from a static
distribution.
This \textit{closed-world assumption} does not hold in real-world settings where
data is continuously drifting~\cite{drift3}.
Consider an image classification task. 
These models assume that: 
(1) the data space is known a priori (\ie, the list of classes is well defined
during training), and 
(2) that the training data is representative of the test data.
These assumptions are invalid in practice due to drift.
This reduces the detection accuracy of these models when drift occurs and the 
distribution of the input data changes.

\PP{Prior Work} 
Recently, researchers have highlighted the challenges associated
with coping with drift~\cite{blazeit,chameleon}.
To detect and recover from drift, the authors recommend that the user
manually identify the evolution of the input distribution and 
construct models specialized for the novel data points (\ie, outliers).
The DBMS then selects the appropriate user-constructed model based on
query-specific accuracy and performance constraints~\cite{blazeit}.
For instance, in case of a traffic surveillance dataset, it uses an 
expensive, more accurate model for object detection under high traffic
conditions and a slower, less accurate model otherwise~\cite{chameleon}.
The key limitation of this approach is that it is not automated.
This delays the drift detection and recovery processes, 
thereby degrading the accuracy and performance of the DBMS.

\PP{Our Approach}
In this paper, we present a visual DBMS, called \sys, that automatically
detects and recovers from drift.
We present an unsupervised algorithm that identifies outliers by learning the
input distribution using adversarial autoencoders.
\sys{'s} \detector relies on a distance metric based on generative adversarial
networks.
We show that this distance metric outperforms state-of-the-art outlier detection
algorithms on high-dimensional visual data (since existing algorithms are tailored for
low-dimensional structured data).
Using \detector, \sys automatically differentiates between key concepts in
the dataset (\eg, weather conditions or time-of-day).
%
%
After detecting drift, \sys{'s} \specializer constructs a family of models
specialized for the novel data points to recover from the changes in input distribution.
We show that the specialized models outperform their non-specialized
counterparts in both accuracy and performance.
We demonstrate that these specialized models are resilient to drift unlike
other forms of model specialization (\eg, student models\cite{teacher}).
Lastly, \sys{'s} \selector picks an ensemble of specialized models from its
family of models for processing a given input.
We compare the efficacy of several model selection policies for drift recovery.
We demonstrate the end-to-end efficacy and efficiency of \sys on high-resolution
dashboard camera videos captured under diverse environments from the Berkeley DeepDrive 
\bdd dataset~\cite{bdd}.

\PP{Contributions}
We make the following contributions:
\squishitemize

\item We present an unsupervised algorithm for drift detection that learns the 
input distribution using adversarial autoencoders.
We propose a novel distance metric based on generative adversarial networks
that is tailored for high-dimensional visual data (\autoref{sec:driftdetection}).

\item We introduce a technique for drift recovery using specialized models
that are resilient to drift. We present a set of policies for selecting an
ensemble of specialized models for processing a given input
(\autoref{sec:specialization}).

\item We implemented our drift detection and recovery algorithms in \sys and 
evaluated its efficacy and efficiency on three datasets. 

\item
\edit{
We demonstrate that \sys delivers 6$\times$ higher throughput and 2$\times$
higher object detection accuracy than a static system without automated drift
detection and recovery.
We show that \sys delivers 1.5$\times$ higher query accuracy than
its static counterpart on canonical aggregation query over visual data
(\autoref{sec:evaluation}). 
}

\squishend

\section{Background}
\label{sec:motivation}

We begin by motivating the need for detecting and recovering from drift
in~\autoref{sec:motivation::example}.
We then present an overview of concept drift to better appreciate the drift
detection and recovery algorithms in~\autoref{sec:preliminaries::drift}.
Lastly, we describe the generative models that \sys uses
in~\autoref{sec:preliminaries::generative}.

\subsection{Motivating Example}
\label{sec:motivation::example}
In this example, we illustrate the benefits of detecting and recovering from
drift by constructing specialized models for novel data points.
\edit{
We compare \sys against a static system with drift detection and recovery
disabled on the \bdd dataset.
This dataset consists of high-resolution, colored images obtained from 
dashboard camera videos under diverse weather conditions~\cite{bdd}.
We examine how a system trained on \rainydata, a cluster in \bdd containing
videos from overcast and rainy days, performs on \daydata, another cluster in
\bdd containing videos from clear, sunny days. 
We defer a detailed description of our experimental setup
to~\autoref{sec:eval::setup}.}

\begin{figure}[t] 
	\centering 
	\includegraphics[width=\linewidth]{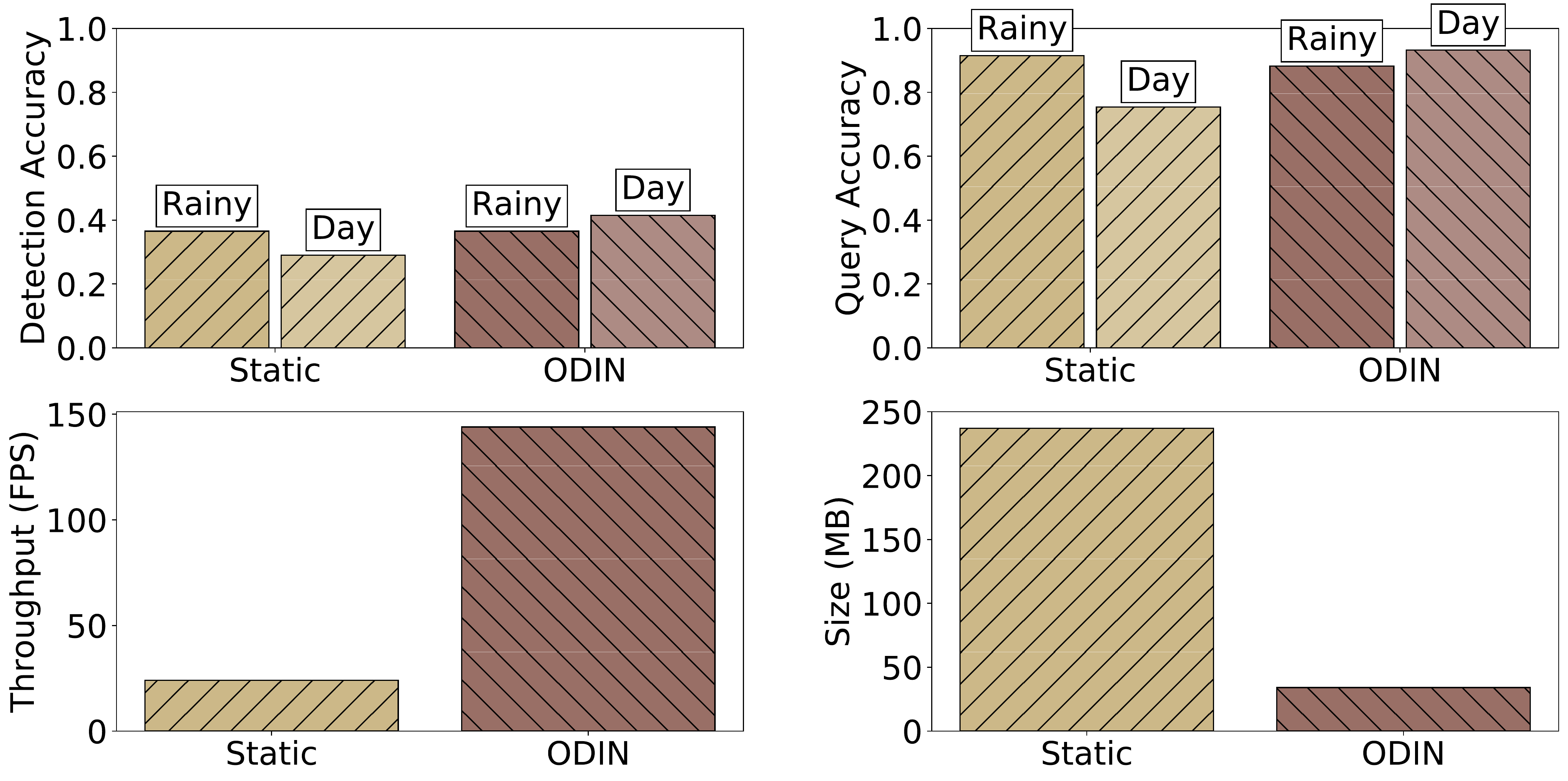}
	\caption{
	\textbf{Motivating Example}: 
	We compare \sys against a static system without automated drift detection and
	recovery on the \bdd dataset.
	}
	\label{fig:motivation}
\end{figure}

\edit{
\sys uses two smaller and faster models specialized for \rainydata and \daydata
clusters. 
In contrast, the static system is a single heavyweight \yolo~\cite{yolo} model
that is trained on \rainydata.
We compare the efficacy and efficiency of the static model against 
the specialized models that are dynamically constructed by \sys after it
detects drift.
The results are shown in \autoref{fig:motivation}.
We compare four metrics:
\squishitemize
\item \textbf{Detection accuracy:} Accuracy of the object detection model.
\item \textbf{Query accuracy:} Accuracy of the output of an aggregation query
counting the number of cars in the videos.
\item \textbf{Throughput:} Number of images processed per second (FPS).
\item \textbf{Memory footprint:} GPU memory occupied by the systems.
\squishend}

\edit{
\sys delivers higher detection and query accuracy than the static system by
leveraging the specialized models for object detection.
The static model trained on the \rainydata subset of \bdd has
lower accuracy when the data changes to \daydata.
\sys automatically detects this drift in the input data and recovers by
deploying a specialized model for \daydata.
So, it maintains higher accuracy even in the presence of drift.
Furthermore, the smaller, specialized models constructed by \sys are 6$\times$
faster and 6$\times$ smaller than the heavyweight model used in the static
system.
We defer a detailed description of the specialized models
to~\autoref{sec:eval::specialization}.
This example illustrates the importance of detecting and recovering from drift.
}

\subsection{Concept Drift}
\label{sec:preliminaries::drift}

\edit{
Concept drift consists of learning in a non-stationary environment, in which
the underlying data distribution (\ie, the joint distribution of the input data
and labels P(X,Y)) evolves over time~\cite{drift2,drift3}.
It is also referred to as \textit{domain adaptation}.
We may classify the changes in the data distribution into two categories:
(1) task drift, and (2) domain drift~\cite{bengio}.
The key distinction between task and domain drift is that the \textit{real}
decision boundary only changes under task drift.}

\edit{
Task drift reflects real changes in the world.
Formally, this corresponds to the drift  in the conditional distribution of
labels given the input data (\ie, P(Y | X)), often resulting from an updated
definition of the task necessitating a change in the predictive function from
the input space to label space.}

\edit{
Domain drift does not occur in reality but rather occurs in the ML model
reflecting this reality.
In practice, this type of drift arises when the model does not identify all the
relevant features or cannot cope with class imbalance.
Formally,  this corresponds to the drift in the marginal distribution of the
input data (\ie, P(X)), with an additional assumption that P(Y | X) remains the
same.}

\edit{
\sys only copes with domain drift.
It measures changes in the marginal distribution of the input data (\ie, P(X)).
\sys uses generative models to construct a low-dimensional
projection of the given images and then clusters the projected images.
We will next provide an overview of generative models.}

\begin{figure}[t]
	\centering
	\begin{subfigure}{.32\textwidth}
		\centering
		\includegraphics[width=\linewidth]{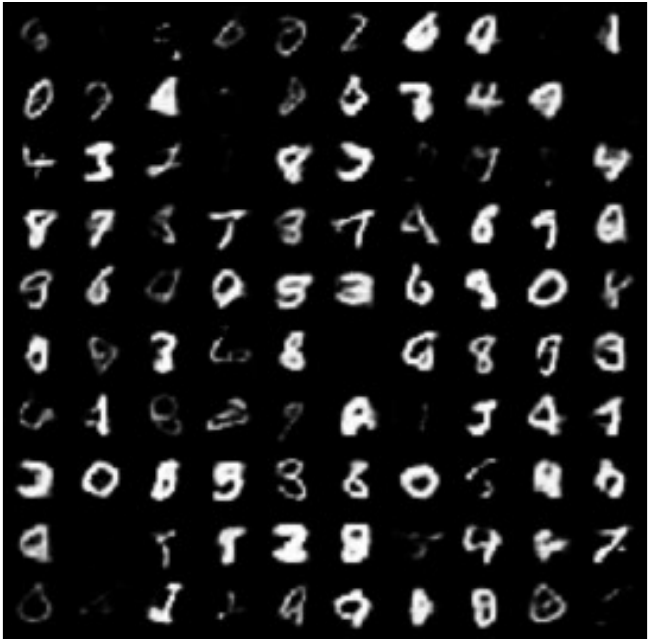}
		\caption{Standard AE}
		\label{fig:aelatent}
	\end{subfigure}
	\hfill
	\begin{subfigure}{.32\textwidth}
		\centering
		\includegraphics[width=\linewidth]{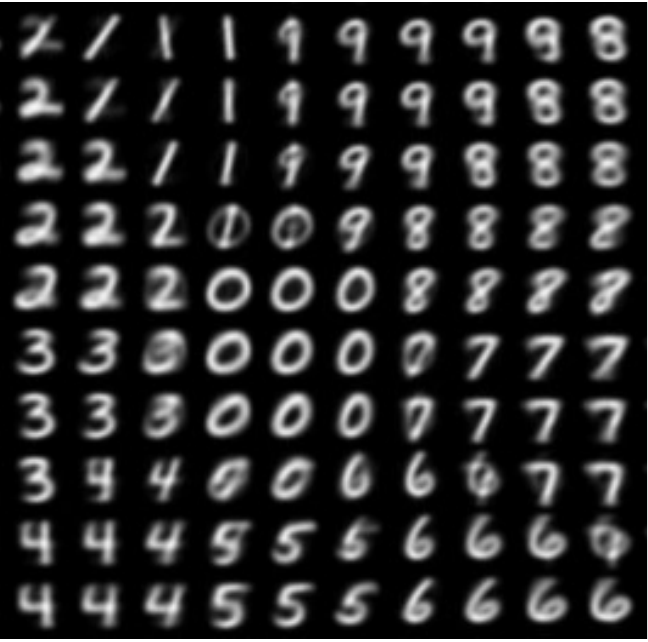}
		\caption{Adversarial AE}
		\label{fig:aaelatent}
	\end{subfigure}
	\hfill
	\begin{subfigure}{.32\textwidth}
		\centering
		\includegraphics[width=\linewidth]{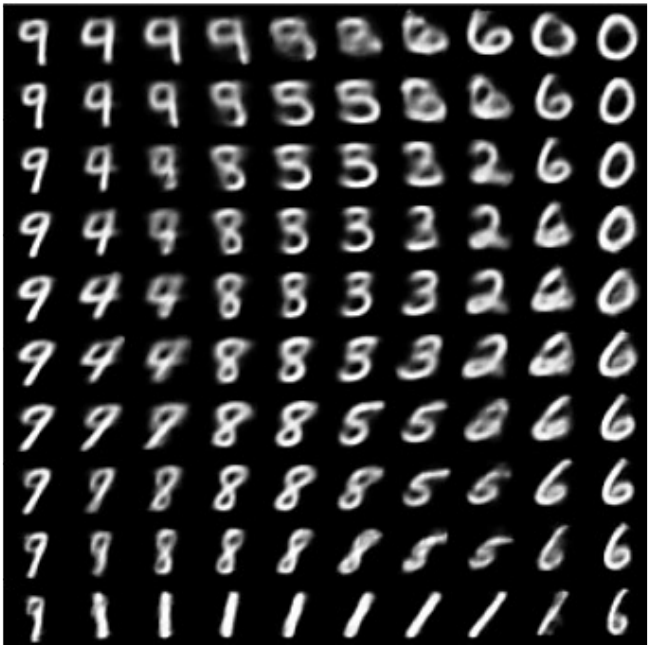}
		\caption{DA-GAN }
		\label{fig:daganlatent}
	\end{subfigure}
	
	\caption{
	\textbf{Latent spaces:} The latent spaces provide crucial clues.
		The standard AE's latent space has holes, indicating unsuitability for drift detection. 
		The adversarial AE's latent space is smooth; the blurriness indicates some loss of information.
		The DA-GAN's latent space is smooth with better reconstruction, indicating most of the underlying distribution
		has been captured in the latent space.}
	\label{fig:latentspaces}
\end{figure}

\subsection{Generative Models}
\label{sec:preliminaries::generative}

Generative models are a category of neural networks for synthesizing new data
points that appear as if they are drawn from the training
data distribution~\cite{gan}.
\sys uses two types of generative models:
(1) autoencoders (AE), and
(2) generative adversarial network (GAN).
Both approaches project an input a low dimensional space by compressing it.
%
%
%
%

%
\sys uses these low-dimensional projections to detect drift by measuring
the distance between existing and novel data points.
Intuitively, GANs and AEs try to capture the most important attributes of images 
during compression, because during training, they must be able to reconstruct an image
from a compressed representation.
Because these low dimensional representations already capture the important attributes,
\ie the underlying distribution, it is easier to detect changes in the underlying distribution
in this space.

\PP{Standard Autoencoder} 
A standard autoencoder (AE)
consists of an encoder and a decoder in series.
%
%
%
%
In an AE:
\squishitemize
\item The encoder $E$ compresses by mapping an input 
image $x$ of $n$ dims to a latent space of $z$ dims, where $n >>
z$.

\item The decoder $G$ takes $z=E(x)$ as input and reconstructs $x$. 
We refer to this reconstruction as $x'=G(E(x))$.
\squishend

An AE is trained using the reconstruction loss 
(binary cross-entropy loss in ~\autoref{eq:recloss}).
AEs display an \textit{irregular mapping} problem 
because of nonlinear activations~\cite{holes}.
It can  project an input to any random point in the latent space $\Re^z$;
this creates holes in the latent space, shown in ~\autoref{fig:aelatent}.
These holes are regions that the decoder cannot reconstruct.
When the underlying distribution changes due to drift, the AE can project 
these new inputs into the holes, leading to empty or invalid reconstructions 
by the decoder.


\PP{Adversarial Autoencoder} 
An adversarial AE closes the holes of the standard AE latent space
by enforcing a smoothness constraint~\cite{aae}
that ensures similar data is projected close together.
This constraint is enforced using a discriminative network $D_Z$.
$D_Z$ takes two inputs: 
(1) points drawn from the latent space, and 
(2) points drawn from a smooth distribution (\eg normal distribution).
It is trained to distinguish between these two distributions using a binary
cross-entropy loss.
In an adversarial AE:
\squishitemize
\item The encoder $E$ maps the input $x$ to a low-dimensional embedding $z$.
Next, $D_Z$ predicts whether $z$ is drawn from the encoded distribution or the 
desired distribution.
\item The decoder $G$ generates $x'$ from $z$. 
The reconstruction loss between $x$ and $x'$ is used to concurrently train
both $E$ and $G$.
\squishend

With the competition between $E$ and $D_Z$, the encoder learns
to map points to the desired distribution, creating a 
latent space without holes (~\autoref{fig:aaelatent}). 
However, the adversarial AE loses some image information, resulting in blurriness.

\PP{Generative Adversarial Network (GAN)} 
%
%
A GAN consists of two networks in series: 
(1) a generator network $G(z)$, and 
(2) a discriminator network $D_I(x)$.
$G(z)$ is similar to the decoder in an AE.
%
%
$D_I(x)$ is similar to the $D_Z(z)$ in an adversarial AE.
%
A GAN uses $D_I(x)$ to improve the quality of the generator.
Given a point $z$ in the latent space, $G(z)$ generates an image $x'$.
Then $D_I(x)$ distinguishes between a real image $x$ and a generated image $x'$.


\PP{Dual Adversarial GAN (DA-GAN)} 
In this paper, we present a novel network that combines the modeling
capabilities of both adversarial AE and GAN.
It consists of four components: 
(1) an encoder, 
(2) a decoder, 
(3) a latent discriminator, and 
(4) an image discriminator.
The decoder of the adversarial AE serves as the generator of the GAN.
The latent and image discriminator together improve the latent space (~\autoref{fig:daganlatent}): 
the latent discriminator makes it smooth, and the image discriminator ensure miinimal information loss
by forcing better reconstruction.
We call this network a \textit{dual-adversarial GAN} because it contains two discriminators.
%
%
We defer a detailed description of DA-GAN to~\autoref{sec:gandistance::dagan}.

\section{System Overview}
\label{sec:approach}

\begin{figure}[t]
	\centering

	\begin{subfigure}{0.9\textwidth}
		\centering
		\includegraphics[width=\textwidth]{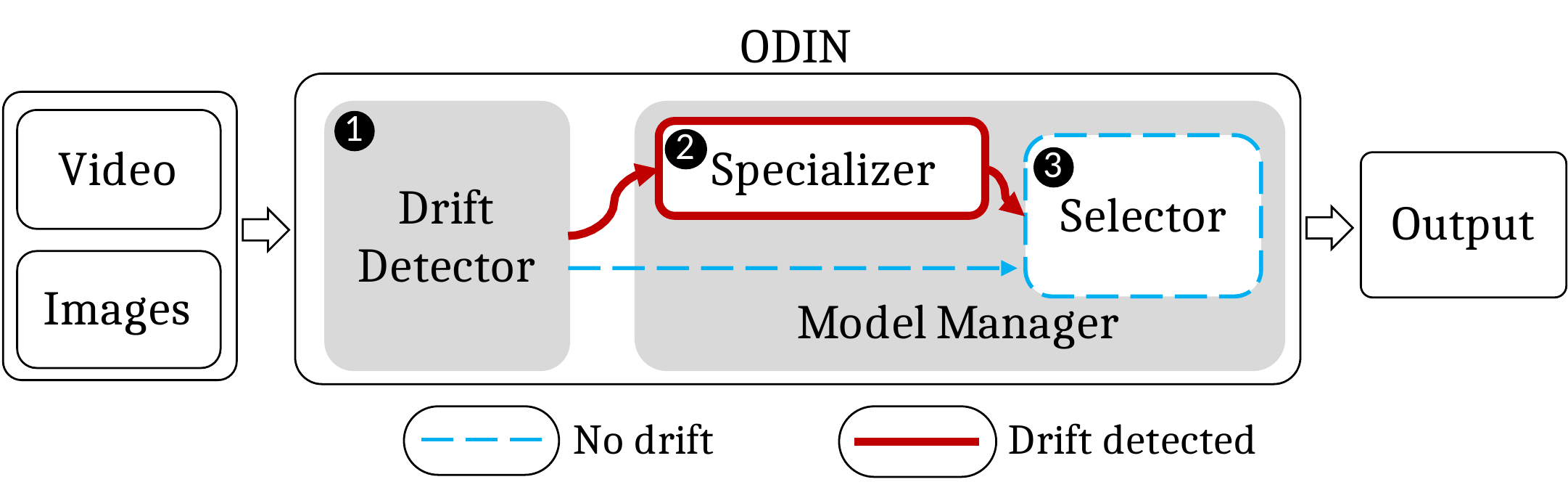}
		\caption{System Architecture}
		\label{fig:system}
	\end{subfigure}\\
	\begin{subfigure}{0.9\textwidth}
		\centering
		\includegraphics[width=\textwidth]{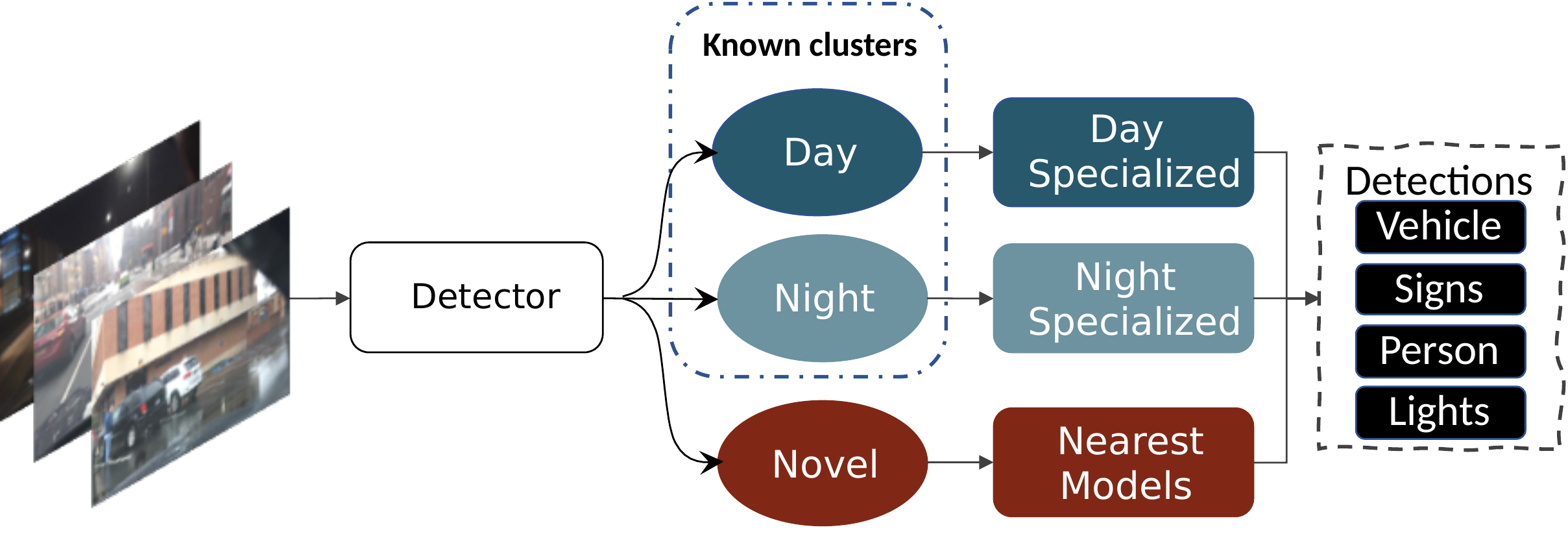}
		\caption{Dataflow}
		\label{fig:dataflow}
	\end{subfigure}

    \caption{\textbf{Architecture and Dataflow of \sys}. 
    \sys takes in a sequence of images as input.
    \dcircle{1} \detector uses a DA-GAN to obtain the low-dimensional
    latent projection of the input and to identify new clusters without 
    supervision.
    \dcircle{2} If drift is detected, \specializer generates a specialized model
    for the newly detected cluster.
    \dcircle{3} Lastly, \selector chooses the appropriate specialized model for
    the given input.
    }
	\label{}
\end{figure}

We now present an overview of the architecture of \sys.
As illustrated in~\autoref{fig:system}, \sys consists of three components: 

\dcircle{1}
\detector identifies drift in the given data using an unsupervised algorithm 
tailored for high-dimensional data.
It learns the distribution of clustered \textit{density bands} in the 
given data.
We present this component in~\autoref{sec:driftdetection}.
Intuitively, a high-density region in the latent space represents a latent
concept and changes in this region indicate changes in the concept itself (\ie,
concept drift).
A key component of \detector is the distance metric that it employs for
clustering data points into density bands in an unsupervised manner.
We show that distance metrics employed for structured data do not work well 
with high-dimensional visual data. 
We make the case for modeling the latent space in visual data using a DA-GAN 
and using its latent space density to detect drift
(\autoref{sec:gandistance::dagan}).

\dcircle{2}
When \detector identifies drift, \sys relies on the \specializer to recover from
the detected drift by generating specialized models for newly detected clusters.
\specializer allows \sys to deliver high accuracy across all clusters.
We present this component in~\autoref{sec:specialization}.
We illustrate the importance of specialization by comparing the accuracy of a 
non-specialized model trained on the entire dataset to specialized models
optimized for particular clusters.

\dcircle{3}
Lastly, the \selector is responsible for choosing the appropriate specialized
model for a given input to perform inference.
When drift occurs, the \specializer may take time to collect sufficient
novel data points before constructing a model for the newly detected cluster.
During this phase, \selector dynamically creates an ensemble of specialized
models from nearby clusters for inference.
We present this component in~\autoref{sec:selector}.
We consider \specializer and \selector to be a part of the \manager within \sys.
\manager is responsible for generating specialized models and choosing the
appropriate one during inference.

\PP{Dataflow} 
\autoref{fig:dataflow} illustrates the flow of data in \sys.
Given an image, \detector performs dimensionality reduction to get its
lower-dimensional manifold.
It uses this manifold to map it to existing clusters from previously seen data. 
If the input belongs to an existing cluster, \selector picks the associated 
model for inference (\eg, identifying objects in the given \bdd image).
If that is not the case, then it picks an ensemble of specialized models from
nearby clusters for inference.
Simultaneously, \specializer records the input to train a specialized
model.

\section{Drift Detection}
\label{sec:driftdetection}

In this section, we present the unsupervised algorithm for drift detection
technique that \sys employs.

\PP{Overview}
\detector performs the following tasks.
\dcircle{1} It first learns a low-dimensional representation of visual data
using DA-GAN (\autoref{sec:gandistance::dagan}).
\dcircle{2} It then uses this low-dimensional representation to cluster the data points 
using without being affected by the curse of
dimensionality\footnote{This phenomena refers to the differencces in classifying high dimensional data vs. 
low dimensional data. Distance metrics tend towards 0 in high dimensions.}.
\dcircle{3} It next constructs a succinct topological representation of these
clusters using \textit{density bands}~\cite{trustclass}
(\autoref{sec:driftdetection::bands}).
This improved representation only captures the \textit{high-density} region of
the cluster, where most of the points exist.
\detector learns the distribution parameters of the density band associated with
 each cluster.
\dcircle{4} Finally, it detects drift by comparing the distribution of novel data
points against that of existing clusters using their KL divergence
(\autoref{sec:driftdetection::bands}).

In the rest of this section, we first formalize the notion of density bands
in~\autoref{sec:driftdetection::bands}. 
We then illustrate the challenges associated with using AEs to detect drift
in~\autoref{sec:gandistance::example}.
We then present how \detector leverages DA-GAN as a distance-preserving
dimensionality reduction technique in high-dimensional spaces
in~\autoref{sec:gandistance::dagan}.
We next discuss how we train DA-GAN in~\autoref{sec:driftdetectiontraining}.
Lastly, we describe how \detector detects drift by comparing the distributions
using KL divergence in~\autoref{sec:gandistance::clustering}.


\subsection{Density Bands}
\label{sec:driftdetection::bands}
Consider a data space $D$.
Let $D_k$ denote the set of points in a cluster $k$ associated with a particular 
concept.
A high-density band in $D_k$ is a subset of $D_k$ that contains more than 50\%
of the points in that cluster.
%

\begin{figure}[t]
	\centering
	\includegraphics[width=0.9\textwidth]{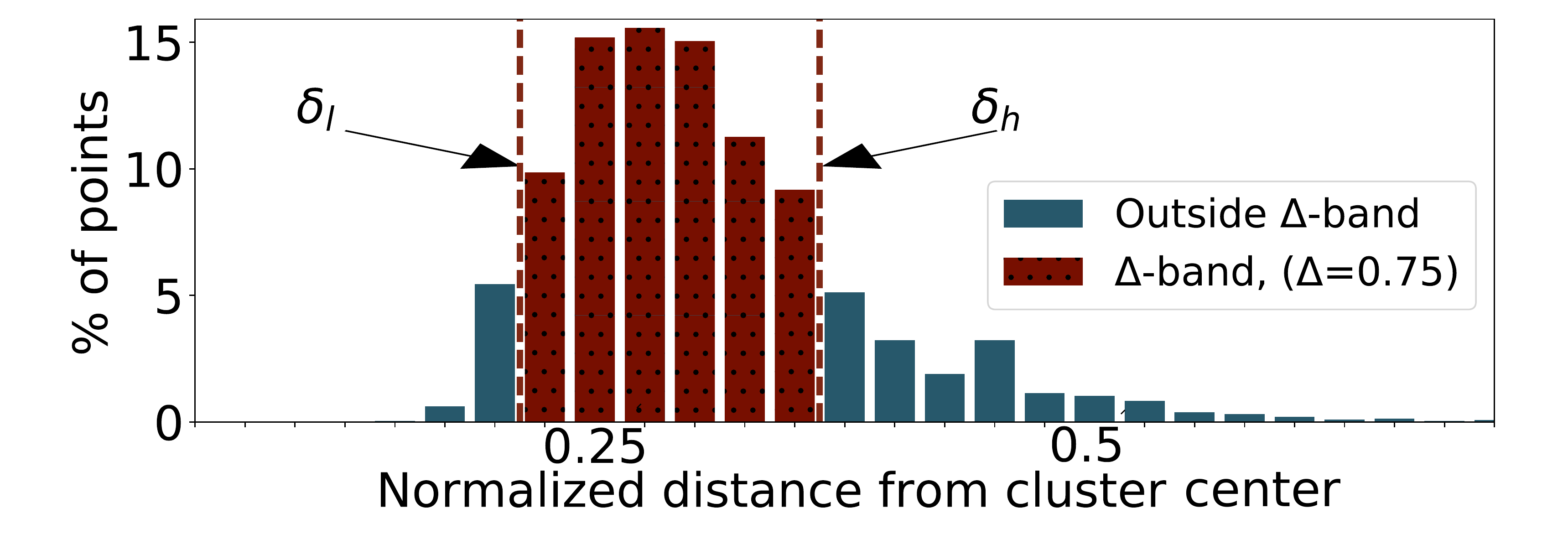}
	\caption{\textbf{Visualization of $\Delta$-band}: 
	Histogram of embedded points in a cluster
	%
	The hypersphere region up to radius $\sim$0.18 is empty. 
	%
	%
	The high-density band, with $\Delta=0.75$, is highlighted, with its 
	bounds $\Delta_l$ and $\Delta_h$.}
	\label{fig:deltaband}
\end{figure}

To obtain a density band, \detector first estimates the distribution of points
in $D_k$.
It centers the band at the distribution peak of the cluster (\ie, where most
points are present with respect to the cluster's center.
It then expands the band inwards to the center and outwards to the cluster
edges to compute the lower and upper bounds of the density band.
For each cluster, \detector uses a pre-defined threshold on the fraction of
points that must be present within the band to determine its bounds; we use
$\Delta=0.5$.
\autoref{fig:deltaband} illustrates this technique for constructing bands.

For a given cluster $D_k$ with $p$ data points, 
\detector computes its centroid $D_k^C= (\sum_{i=1}^{p} x_i)/ p $ and its
probability mass function.
Let $f_{\Delta}(x)$ be a continuous density function of $D_k$ that is
estimated on a normalized distance metric $d:\Re^n\rightarrow[0,1]$.
Here, $d$ measures the distance between any point $x_i$ and the centroid
$D_k^C$. 
Thus, $f_{\Delta}(x)$ captures the distribution of $D_k$'s points with respect to the
centroid $D_k^C$.
\detector then uses $f_{\Delta}(x)$ to compute the density band.
A density band $\Delta$ is defined by two bounds: $[\Delta_l,\Delta_h]$,
where $0\leq\Delta_l <\Delta_h\leq1$. 
As shown in~\autoref{fig:deltaband}, 
$\Delta$ represents the fraction of points in the cluster within the lower and
upper bounds $\Delta_l$ and $\Delta_h$, respectively.
\detector computes the density band's parameters based on $\Delta$ and its
density function:

\begin{equation}
\label{eq:dband}
\int_{\Delta_l}^{\Delta_h}f_{\Delta}(x)dx=\Delta
\end{equation}

\PP{KL Divergence} 
While processing a given data point, \detector maps it to existing permanent
clusters associated with known concepts or to a single temporary cluster.
We defer a detailed description of this algorithm to~\autoref{sec:specializer}.
A point that falls inside a permanent cluster's $\Delta$-band is assigned to 
that cluster.
A point that falls outside all of the permanent clusters' $\Delta$-bands is
assigned to the temporary cluster.
\detector continuously updates the parameters of the temporary cluster's
$\Delta$-band based on the new data points in the input stream.
It detects drift by using KL divergence to compare the posterior distribution 
of the temporary cluster's $\Delta$-band after a point is added against the
prior distribution before a point is added.
The KL divergence between the two distributions modeling a data point $x$
(\ie, the prior $P_A$ and the posterior $P_B$) is given by:

\begin{equation}
D_{KL}(P_A||P_B)=-\sum_{x\in X}P_A(x)\log(P_B(x)/P_A(x))
\label{eq:klmetric}
\end{equation} 

Here $P_A$ is the \textit{expected} prior and $P_B$ is
the \textit{live} posterior observed in practice.
When  the distribution inside the $\Delta$-band before and
after a point is added no longer changes ($D_{KL}\rightarrow 0$ when $P_B$ $=$
$P_A$), \sys consider the temporary cluster as stable.
This indicates the presence of drift, since there are enough points 
in the temporary cluster to indicate the introduction of a new concept (\eg,
snowy images).
\sys converts the temporary cluster to a permanent cluster and adds it to the 
set of permanent clusters (\autoref{sec:selector}).
It concurrently constructs a specialized model for this cluster
(\autoref{sec:specializer}).
Lastly, it initializes a new empty temporary cluster for processing subsequent
points.

\PP{Manifold Learning}
With KL divergence, \detector measures the changes in the input distribution. 
However, it still needs a suitable distance metric for modeling the distribution
of the input and for projecting it from high-dimensional images to a
lower-dimensional manifold.
We next illustrate the challenges associated with detecting drift in
high-dimensional visual data.

\subsection{Drift Detection in Images}
\label{sec:gandistance::example}

Real-world datasets often fit a manifold whose dimensionality
is lower than the raw data.
Consider the digit-classification task on the \mnist dataset~\cite{mnist}.
We may project the $28\times 28$ images in this dataset (\ie, 784 dimensions)
onto a ten-dimensional manifold of digits using a neural network with one-hot
encoded outputs.
Here the network $N_{MNIST}:\Re^{784} \rightarrow\Re^{10}$ learns to approximate
the projection from the raw data to the manifold.
While this network works well in the absence of concept drift, 
it will start to misclassify novel data points in the presence of drift.
This is because the changes in the data distribution necessitate a shift in the
projection as well.

\begin{figure}[t] 
    \centering \includegraphics[width=0.9\textwidth]{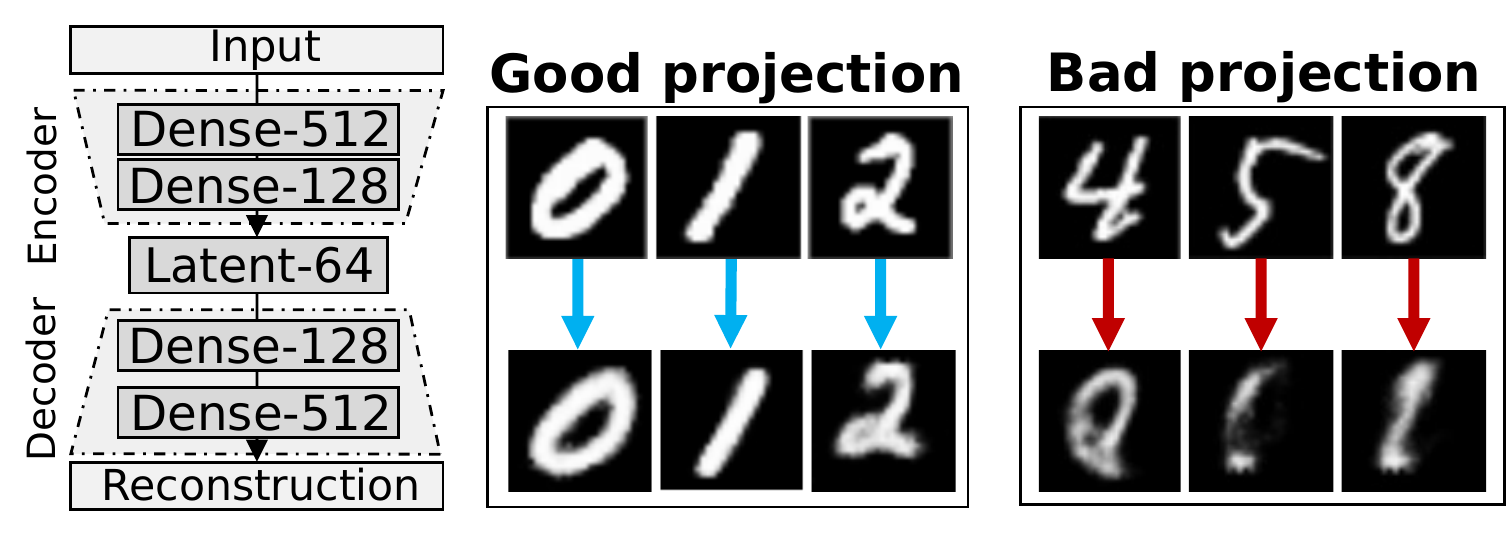}
	\caption{\textbf{Projection Failure}: We use two similar datasets to 
	demonstrate the projection failure in the presence of concept drift.
	 A model trained on a subset of \mnist (digits \ttt{0--2}) fails to reconstruct
	 outliers (digits \ttt{3--9}).
	 }
	\label{fig:driftae}
\end{figure}

\autoref{fig:driftae} illustrates this problem of a \textit{projection failure}. 
We train an AE with four dense layers each with ReLU activation. 
The dimensionality of the latent space of this AE is 64. 
We train it on a subset of \mnist containing three digits \ttt{0--2} and then
test it on the entire dataset.
We observe that the AE cannot reconstruct digits \ttt{3--9}, 
since it expects the images to be drawn from a distribution comprising of 
digits \ttt{0-2}.
Even though the images are visually similar (\ie, $28\times 28$ black-and-white
images of digits), the concept drift in the testing inputs causes the
reconstruction to fail.
This is because the AE only learns the projection of digits \ttt{0--2}, 
instead of learning the projection of black-and-white images in general.

The most notable observation from this experiment is that high reconstruction
error of the AE indicates drift.
This means the latent space for drifted data is far from the latent space of
known data, since the autoencoder only learns the projection over the known data.
Since the latent space is at a lower dimensionality than the input images,
measuring drift there bypasses the curse of dimensionality and can be more effective.
So, \detector measures drift using the latent space representation.
However, AEs have problems in latent space representation, such as 
holes(\autoref{fig:aelatent}).
%
%
We next present a distance-preserving dimensionality reduction technique that
works well on high-dimensional images and avoids problems of AEs.

\subsection{Dual-Adversarial GAN}
\label{sec:gandistance::dagan}
%
\detector computes $\Delta$-bands and KL divergence on this latent space.
As we discussed in the overview of generative models
(\autoref{sec:preliminaries::generative}):
(1) AEs create holes in the latent space during projection,
(2) Adversarial AEs lose some information in the image while constructing
smoother projections, and
(3) GANs are designed  for image synthesis, not representation modeling.

\PP{Overview}
We present a network, called Dual-Adversarial GAN (DA-GAN), that combines an
adversarial AE and a GAN to exploit their latent encoding and image information
preserving properties, respectively.
We use this network to map images to a low-dimensional latent space.
The adversarial AE ensures that the latent space matches the desired smooth
distribution (\eg, normal distribution).
The GAN ensures that the latent space does not lose important information during
encoding by focusing on image reconstruction.
Since the latent discriminator is trained on the desired smooth distribution, 
it is adept at discriminating the inlier frames from the outlier frames which
should be mapped to a different distribution~\cite{drae}.
In this manner, DA-GAN functions as a distance-preserving projection technique
that works well on high-dimensional data.

\begin{figure}[t]
	\centering
	\includegraphics[width=0.9\textwidth]{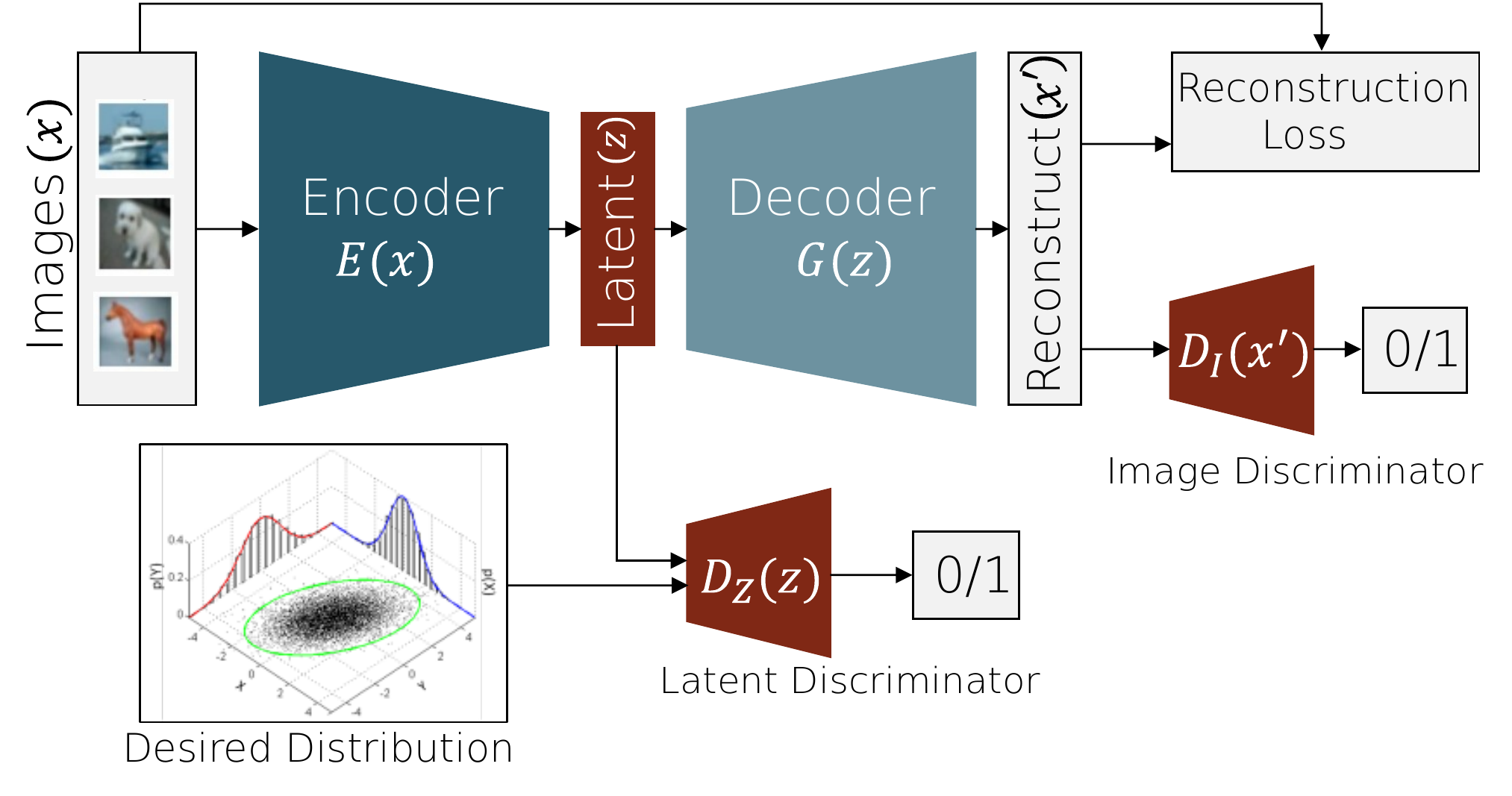}
	\caption{\textbf{Dual-Adversarial GAN}: 
	The dual adversarial GAN has three loss functions: 
	(1) latent discriminator loss ($\mathcal{L}_{Z}$), 
	(2) image discriminator loss ($\mathcal{L}_{I}$), and 
	(3) reconstruction loss ($\mathcal{L}_{R})$.}
	\label{fig:dagan}
\end{figure}

\PP{Structure}
\label{sec:gandistance::dagan:::dagan-losses}
The structure of the DA-GAN is shown in~\autoref{fig:dagan}.
It consists of four components:
\dcircle{1} Encoder $E(x)$,
\dcircle{2} Decoder $G(z)$,
\dcircle{3} Latent discriminator $D_Z(z)$, and 
\dcircle{4} Image discriminator $D_I(x)$.
We keep the basic structure of an autoencoder and a GAN intact 
by using an encoder and a decoder .
The encoder maps an input $x$ to the latent space: $z=E(x)$
The decoder seeks to reconstruct $x$ using $z$: $x'=G(z)$.
We introduce two adversarial discriminators. 

\dcircle{3}
The first discriminator $D_Z(z)$ imposes a prior on the latent space $z$.
$D_Z(z)$ learns to minimize a binary cross-entropy loss $\mathcal{L}_Z$ 
between points drawn from the
normal distribution $\mathcal{N}(0,1)$ and from the encoded latent space
$D_Z(E(x))$:
%

\begin{equation}
\mathcal{L}_Z=\log(D_Z(\mathcal{N}(0,1))) + \log(1-D_Z(E(x))) 
\label{eq:latentdiscriloss}
\end{equation}
	
$D_Z(z)$ forces the encoder $E(x)$ to learn clearer separations between classes
and to create better reconstructions.

\dcircle{4}
The second discriminator $D_I(x)$ counters the blurriness caused by loss of
information in the latent space~\cite{pixelgan}.
This adversarial image discriminator operates on the output of the decoder
$G(z)$ (\ie, $x'$).
It learns to minimize  $\mathcal{L}_{I}$
in~\autoref{eq:advrecloss}.
$\mathcal{L}_{I}$ is a binary cross-entropy loss to compare the true image $x$ and a
reconstruction from a random point in the normal distribution
$G(\mathcal{N}(0,1))$. 

\begin{equation}
\mathcal{L}_{I}=\log(D_I(x)) + \log(1-D_I(G(\mathcal{N}(0,1))))
\label{eq:advrecloss}
\end{equation}

$D_I(x)$ reduces information loss by forcing $E(x)$ to encode more useful
information, allowing $G(z)$ to create better
reconstructions.

Lastly, we use the pixel-wise reconstruction loss $\mathcal{L}_{R}$ 
in~\autoref{eq:recloss} to compare the input image $x$ to the output image
$x'=G(E(x))$.

\begin{equation}
\mathcal{L}_{R}=-E_z[\log(x'|x)]
\label{eq:recloss}
\end{equation}

The two adversarial losses $\mathcal{L}_{Z}$ and $\mathcal{L}_{I}$ impose the
dual constraints of:
(1) smoother latent space without holes, and
(2) high quality encoding with minimal loss of information, respectively.
%

\begin{figure}[t] 
    \centering 
    \includegraphics[width=0.9\textwidth]{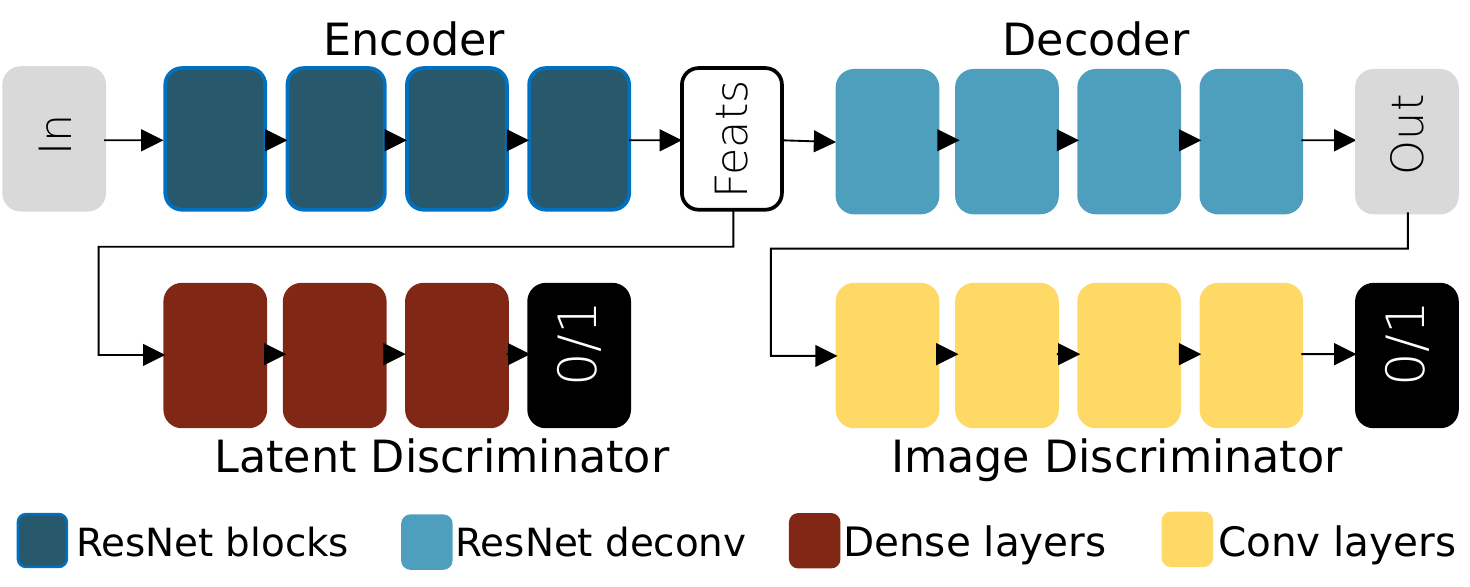}
	\caption{\textbf{DA-GAN details}: 
	The blocks of the DA-GAN model. 
	The encoder and decoder are derived from Resnet \protect\cite{resnet}. 
	The output of the encoder consists of 1024 features.
	We illustrate the adversarial discriminators that contribute 
	to $\mathcal{L}_Z$ and $\mathcal{L}_I$.}
	\label{fig:dagannetwork}
\end{figure}

\PP{Construction}
\label{sec:gandistance::dagan:::daganconstruction}
~\autoref{fig:dagannetwork} illustrates the components of DA-GAN.
The encoder consists of Resnet blocks\cite{resnet}.
We pool the final features by channel to extract the 
distribution features representing the input. 
During tranining, these features are passed to the latent discriminator.
%
%
The distribution features are also passed through deconvolutional Resnet
blocks to reconstruct the original input, and the reconstructed image 
is passed through the image discriminator. 
%

\PP{Comparison to U-NET}
U-NET is a neural network that performs residual transfer from encoder to
decoder by bypassing the latent space~\cite{unet},
allowing U-NET to learn to better reconstruct the image.
However, bypassing the latent space skips information from 
being encoded in the latent space.
This would prevent drift detection, since the underlying distribution
is not properly encoded.

\subsection{DA-GAN Training}
\label{sec:driftdetectiontraining}
We partition each dataset into two sets of classes:
(1) \textit{known classes} for training the \detector, and
(2) \textit{unknown classes} for testing the \detector.
The training procedure ensures that DA-GAN learns to model the distribution of
the known classes.
During testing, we present data points from both known and unknown classes to 
evaluate the ability of the \detector to detect drift.

\PP{Loss Functions}
The overall loss function for the GAN is shown in~\autoref{eq:ganeq}.
We compute the weighted sum of three loss functions:
(1) the latent discriminator loss ($\mathcal{L}_{Z}$), 
(2) the image discriminator loss ($\mathcal{L}_{I}$), and
(3) the standard reconstruction loss ($\mathcal{L}_{R}$). 

\begin{equation}
\mathcal{L}=\lambda_Z\mathcal{L}_Z + \lambda_I\mathcal{L}_I +
\lambda_R\mathcal{L}_{R}
\label{eq:ganeq}
\end{equation}

Since the discriminators are adversarial, they must be equally weighted.
If $\lambda_Z$ and $\lambda_I$ differ, then the learning landscape is more
difficult and render the training procedure to be unstable.
Thus, we let $\lambda_Z=\lambda_I=1$. 

The reconstruction loss $\lambda_R$ ensures the encoder is encoding enough
information in the latent space for the decoder. 
We do not require the synthetic images created by the decoder. 
We only need the decoder to be good at reconstructing the input, since that
demonstrates that the encoder is recording relevant information in the latent
space.
However, $\lambda_R$ also creates the holes in the latent space, as shown
in~\autoref{fig:aelatent}.
We ensure that the latent discriminator loss is prioritized over $\lambda_R$ 
by setting $\lambda_R=0.5\lambda_Z=0.5$, which closes the latent space holes.

\PP{Training}
We use the adversarial training procedure shown in~\autoref{alg:training}.
In each iteration, the components of DA-GAN are updated sequentially.
The image discriminator is trained to distinguish between real images
and synthetic images generated by the decoder (Lines 5-7).
The decoder is trained to fool the image discriminator so that it mistakes
synthetic images for real images (Line 8).
The latent discriminator is trained to distinguish between points drawn from a
normal distribution and points generated by the encoder (Lines 9-11).
The encoder is trained to fool the latent discriminator so that it mistakes the
points generated by the encoder for points coming from a normal distribution
(Line 12).
The encoder will succeed only if it maps the input images to the desired smooth 
distribution.
Finally, the autoencoder is updated to minimize the pixel-wise reconstruction
loss (Lines 13).
This training procedure allows DA-GAN to deliver high image fidelity, 
since both encoder and decoder must work together to reconstruct the input.

\begin{algorithm}[t]
	\caption{\textbf{DA-GAN Training Iteration}}\label{alg:training}
	\scriptsize
	\DontPrintSemicolon
	\SetNoFillComment
	\SetKwInOut{Input}{input}
	\SetKwInOut{Output}{output}
	\SetKwInOut{Parameters}{parameters}
    \SetKwInOut{Functions}{functions}
	\SetAlgoLined
	\Input{Encoder \ttt{E()}, Decoder \ttt{G()}, Latent Discriminator $\mathtt{D_z()}$, Image Discriminator $\mathtt{D_I()}$}
	\Output{Trained DA-GAN}
    \Functions{\ttt{BCE(a,b)}: Binary cross entropy loss between $a,b$; \ttt{GetRandomNormal()}: Sample numbers from $\mathcal{N}(0,1)$;  \ttt{GetRandomBatch()}: Sample images from \bdd;
    \ttt{Backpropagate(a)}: Backpropagate loss $a$ over DA-GAN}
	\BlankLine
        \tcp{Set up targets}
        $y_{real},\,\,y_{fake} = \mathtt{ones(}\mathtt{)},\,\,\mathtt{zeros(}\mathtt{)}$\;
        $z_{real},\,\,z_{fake} = \mathtt{ones(}\mathtt{)},\,\,\mathtt{zeros(}\mathtt{)}$\;
        \tcp{Get Minibatches}
        $z'\leftarrow \mathtt{GetRandomNormal}()$;\,\,\,\, $x\leftarrow \mathtt{GetRandomBatch}()$\;
        $x'=\mathtt{G}(z')$;\,\,\,\, $z=\mathtt{E}(x)$\;
        \tcp{Update the Image Discriminator}
        $\mathtt{LossReal\_D_I}=\mathtt{BCE}(\mathtt{D_I}(x), y_{real})$\;
        $\mathtt{LossFake\_D_I}=\mathtt{BCE}(\mathtt{D_I}(x'), y_{fake})$\;
        $\mathtt{Backpropagate}(\mathtt{LossReal\_D_I}+\mathtt{LossFake\_D_I})$\;
        \BlankLine
        \tcp{Update the Decoder}
        $\mathtt{Backpropagate}(\mathtt{BCE}(\mathtt{D\_I}(x'),y_{real}))$\;
        \BlankLine
        \tcp{Update the Latent Discriminator}
        $\mathtt{LossReal\_D_z}=\mathtt{BCE}(\mathtt{D_z}(z'), z_{real})$\;
        $\mathtt{LossFake\_D_z}=\mathtt{BCE}(\mathtt{D_z}(z), z_{fake})$\;
        $\mathtt{Backpropagate}(\mathtt{LossReal\_D_z}+\mathtt{LossFake\_D_z})$\;
        \BlankLine
        \tcp{Update the Encoder}
        $\mathtt{Backpropagate}(\mathtt{BCE}(\mathtt{D_z}(z), z_{real}))$\;
        \BlankLine
        \tcp{Update both Encoder and Decoder}
        $\mathtt{Backpropagate}(0.5\cdot \mathtt{BCE}(x,x'))$
\end{algorithm}

\subsection{Clustering}
\label{sec:gandistance::clustering}

Lastly, we describe how \detector detects drift using DA-GAN
(\autoref{sec:gandistance::dagan}) and $\Delta$-bands with KL divergence
(\autoref{sec:driftdetection::bands}).
\detector first projects images to a lower dimensional manifold using DA-GAN.
After training DA-GAN, \detector only uses the encoder for projecting images.

While processing a stream of incoming data points, \detector maintains a
collection of clusters.
For a given point, \detector first projects it to a low-dimensional manifold
using DA-GAN's encoder.
\dcircle{1}
If the point falls within the $\Delta$-band of an existing cluster, 
it is added to that cluster.
\detector updates that cluster's $\Delta$-band using~\autoref{eq:dband}.
\dcircle{2}
If the point falls outside all of the existing $\Delta$-bands, 
then \detector adds it to a temporary cluster.
It recomputes the temporary cluster's $\Delta$-band and distribution.
When that $\Delta$-band's upper and lower bounds no longer change and the
distribution stabilizes as per~\autoref{eq:klmetric}, 
\detector converts the temporary cluster to a permanent cluster and adds it to
its collection of clusters.
It then initializes a new empty temporary cluster to process subsequent points.
The addition of a cluster indicates drift 
(\ie, the discovery of a new region in the input data space).

All of the components of \detector work in tandem to circumvent the curse of
dimensionality (~\autoref{sec:driftdetection}).
For instance, \bdd{'s} $1280 \times 720$ colored camera images contain
$\sim$921K dimensions.
DA-GAN's encoder projects these high-dimensional images down to 1024 dimensions 
(~\autoref{fig:dagannetwork}) while generating clusters.
\detector then maps these clusters to $\Delta$-bands with four dimensions:
(1) lower bound $\Delta_l$, 
(2) upper bound $\Delta_h$, 
(3) prior $P_A$, and
(4) posterior $P_B$.
Lastly, it detects drift using these $\Delta$-bands and KL divergence.
In this manner, we reduce the dimensionality of the drift detection 
problem from $\sim$921K dimensions to four dimensions.
We demonstrate the efficacy of \sys{'s} drift detector on diverse
datasets in~\autoref{sec:eval::detection}.

\section{Drift Recovery}
\label{sec:specialization}

In this section, we discuss how \sys recovers from drift.
When \detector creates a new cluster after identifying drift, 
\specializer generates a new model that is tailored for the newly
detected cluster (\autoref{sec:specializer}).
We illustrate the types of models that \specializer constructs through a case
study on object detection (\autoref{sec:selector}).
Lastly, after constructing the models, \manager uses \selector to pick the
best-fit specialized models for prediction (\autoref{sec:selector}).

\subsection{Model Specialization}
\label{sec:specializer}

A model $\mathcal{M}_{k}$ tailored for a cluster $D_{k}$ learns a mapping from
that cluster's data points to labels $\mathcal{Y}$.
\manager maintains a collection of models:

\begin{equation}
\{\mathcal{M}\}^n:\{D\}^n\rightarrow\mathcal{Y}
\end{equation}

Here, $n$ represents the currently materialized set of models. 
When \detector identifies a new cluster $D_{k+1}$, 
\specializer constructs a model $\mathcal{M}_{k+1}$ optimized for the points in
$D_{k+1}$ and adds it to the set of materialized models.
\detector typically maps most of the data points to existing clusters.
For these inliers, \sys only updates their corresponding model with the
new data in the associated cluster.
\specializer constructs new models only to cope with the outliers.

\begin{algorithm}[t]
	\caption{\textbf{Model Specialization}}\label{alg:virtualdrift}
	\scriptsize
	\DontPrintSemicolon
	\SetNoFillComment
	\SetKwInOut{Input}{input}
	\SetKwInOut{Output}{output}
	\SetKwInOut{Parameter}{parameter}
	\SetAlgoLined
	\Input{$n$ models $\{\mathcal{M}\}^n$, data point $x_i$, computer vision task
	$\mathcal{T}$}
	\Output{$\mathcal{T}(x_i)$, updated models and clusters if needed}
	\Parameter{$d$ (\texttt{DA}-\texttt{GAN} distance metric)}
	\BlankLine
	$\texttt{cluster\_found} = \texttt{False}$\;
	\For{$\mathcal{M}_j \in \{\mathcal{M}\}^n$}{
		\tcp{Distance to centroid using DA-GAN}
		$d'_{x_i} = d(x_i, \mathcal{M}_{j}^{centroid})$\;
		\tcp{Check if inside $\Delta$-band of $\mathcal{M}_j$}
		\If{$\Delta_l^j < d'_{x_i} < \Delta_h^j$}{
			\tcp{Add point to model's data $D_{j}$}
			$D_{j} = D_{j} \cup x_i$\;
			\tcp{Update the parameters}
			$\mathtt{UpdateDeltaBand(}D_{j}\mathtt{)}$;\,\,\,\,$\mathtt{UpdateModel(}\mathcal{M}_j\mathtt{)}$\;
			\tcp{Flag for found cluster}
			$\texttt{cluster\_found} = \texttt{True}$\;
	}}
	\If{$\texttt{cluster\_found} = \texttt{False}$}{
		$D_G = D_G\cup x_i$\;
		\tcp{Update the distributions}
		$\mathtt{UpdateDeltaBand(}D_G\mathtt{)}$\;
		\If{$\mathtt{StableDistribution(}D_G\mathtt{)}$}{
			$\mathcal{M}_{n+1}\leftarrow\mathtt{GenerateNewModel(}D_G\mathtt{)}$\;
			$D_{n+1}\leftarrow\mathtt{GenerateNewCluster(}D_G\mathtt{)}$\;
		}
	}
\end{algorithm}

\PP{Model Generation}
~\autoref{alg:virtualdrift} presents the algorithm for generating models.
\sys uses the distance metric (in this case, DA-GAN) to determine whether
specialization is necessary.
For each input $x_i$, the DA-GAN projects it to the latent space. 
For each existing cluster generated by \detector, 
\specializer checks if $x_i$ belongs to that cluster by comparing the distance
between the projected $x_i$ and the center of the cluster (Line 3) 
against the lower and upper bounds of that cluster's $\Delta$-band (Line 4). 
If the point exists in a cluster, 
\specializer updates the cluster's distribution parameters and model 
using $x_i$ (Lines 5-6; lower dashed line in \sys{'s} system design in
~\autoref{fig:system}).
If $x_i$ belongs in no existing cluster, then \sys uses \detector to add it 
to the temporary cluster $D_G$ (Line 11).
%
%
Under the gradual drift assumption, $D_G$ grows over time as new outliers are
added.
When $D_G$'s $\Delta$-band no longer exhibits changes (Line 13), 
\specializer constructs a new model trained on $D_G$ (Line 14) and
\detector creates a new cluster (Line 15).

\subsection{Types of Specialized Models}
\label{sec:yolo}
\sys adopts two approaches to model specialization:
\squishitemize
\item \textbf{Specialized models for improved accuracy}: 
Upon discovering a new cluster $D_k$ in the data space, \specializer 
generates a \textit{specialized} model $\mathcal{M}_k$ to perform the given task 
on $D_k$.
%
%

\item \textbf{Lite models for improved performance}: 
\specializer uses a student-teacher
approach to train a faster, weaker student model(called \yololite) using the outputs of the slower
parent model~\cite{teacher}.
%
%
\squishend

\PP{Specialized vs Lite Models}
Lite models sacrifice accuracy to enable faster training and subsequent
deployment.
This is because, unlike specialized models, they do not require \textit{oracle
labels} from humans or weakly-supervised agents~\cite{snorkel} during training.
They instead leverage the outputs of an existing parent model~\cite{teacher}.
Thus, \sys trains and deploys a lite model as soon as it detects a new cluster.
Later, when the oracle labels for the newly detected cluster is available, 
\specializer trains a specialized model and replaces the lite model with its
specialized counterpart.
We examine the efficacy and efficiency of these two types of models
in~\autoref{sec:eval::specialization}.

\PP{Case Study: Object Detection}
We next illustrate how \sys specializes models for recovering from drift
through a case study on object detection using the \yolo model~\cite{yolo}.

\PP{YOLOv3} 
\sys uses the \yolo object detection model as the baseline object detector
($\mathcal{M}$).
\yolo is an efficient detector that performs inferences in a single pass over
the images.
It generates region proposals and combines them with a classification model to
concurrently segment images and perform dense object labeling.

The \yolo network consists of 24 convolutional layers and 2
fully-connected layers.
It divides the input image into a $s\times s$ grid with $k$ bounding boxes per
grid.
It assigns a \textit{confidence score} for each bounding box: 
$\mathcal{C} = \texttt{P}(obj)\cdot \texttt{IOU}(true, pred)$.
Here, $\texttt{P}(obj)$ is the probability of an object in the bounding box
and $\texttt{IOU}$ is the intersection over union of the true bounding box and the
predicted bounding box.
It also predicts a \textit{class probability} for each bounding box.
We train the  \yolo model by: 
(1) minimizing the bounding box prediction error to ensure
$\texttt{IOU}(true,pred)\rightarrow 1$, and
(2) maximizing the probability of correct class predictions.

While \yolo is accurate on challenging datasets, it is computationally
expensive and requires multiple GPUs for real-time operation (\eg, 40~fps). 
Furthermore, the model is designed for dense, generalized object detection on
the COCO dataset that contains a wide array of classes~\cite{cocoapi}.
The resultant model complexity is not justified when it is geared towards a
particular domain with a narrow set of classes 
(\eg, dashboard camera videos in \bdd). 
In this scenario, specialized models deliver higher performance.

\PP{YOLO-Specialized}
To construct a specialized \yolo model that is specialized to one cluster, 
we first prune a subset of
convolutional layers from the original model while preserving sufficient
accuracy on the given task and dataset.
The resulting model, which we refer to as \yololocal, is capable of
object detection with fewer computational resources and a smaller memory
footprint. 
\specializer trains the \yololocal model on the data points in a particular
cluster.
\sys builds specialized models for each detected cluster. 
Since it optimizes these models for a subset of the data space, they are smaller
and support faster inference.
Unlike the baseline network, \yololocal only contains 9 convolutional layers.
Since these models are smaller, they do not suffer from the vanishing gradient
problem during training.
Thus, we remove the batch normalization layer from the network.

\PP{YOLO-Lite}
To construct a lite \yolo model, we use the \yololocal model architecture 
and train it with the outputs of the original \yolo model.
We refer to this lite model as \yololite.
This model approximates its teacher's accuracy (\ie, \yolo) at higher
throughput.
Compared to \yololocal, \yololite is easier to train since it does not require
externally sourced oracle labels.
\specializer directly use the outputs of the \yolo model on the newly detected
cluster to train a \yololite model.

\subsection{Model Selection}
\label{sec:selector}

After \detector and \specializer have identified the new clusters and generated
specialized models, \sys relies on the \selector to pick the
appropriate specialized models for prediction.
Typically, for a given point, \selector chooses the specialized model associated
with that point's cluster.
However, in the presence of drift, \detector is actively assigning points to a
new cluster and \specializer is yet to construct a model for that cluster.
In this scenario, \selector must choose amongst the existing specialized models
associated with closely-related clusters.

\selector employs a model ensemble selection policy: 
$S_k:x_i\rightarrow \{M\}^k$, to pick $k$ best-fit models to operate on $x_i$.
We examine the following selection policies:
\squishitemize
\item \textbf{$k$-nearest models: unweighted (KNN-U)}.
Under this policy, \selector picks the $k$ nearest models based on distance
between $x_i$ and the cluster centroids.

\item \textbf{$k$-nearest models: weighted (KNN-W)}. 
Under this policy, \selector picks the same set of models as the prior policy.
However, it prioritizes these models based on the distances $\{d\}^k$ between
their clusters' centroids and $x_i$. 
The weights are inversely proportional to distance (\ie, the closest cluster
gets the highest priority):
\begin{equation}
w_m = d'_i/\sum d'_i
\label{eq:weighting}
\end{equation}

Here, $w_m$ is the weight of the model associated with cluster $m$ and $d'_i =
\mathtt{max(}\{d\}^k\mathtt{)}/d_i$ is the inverted distance.

\item \textbf{$\Delta$-band models ($\Delta$-BM)}. 
Under this policy, \selector picks the models associated with all of the
clusters whose $\Delta$-bands contain $x_i$.
If $x_i$ does not fall within any $\Delta$-band, then \selector falls back to
the KNN-W policy.

\squishend

\section{Evaluation}
\label{sec:evaluation}

Our evaluation of \sys aims to answer the following questions:

\squishitemize
\item  
Is \detector effective at identifying drift compared to the state-of-the-art
outlier detection algorithms? (\autoref{sec:eval::detection})

\item 
Are the models constructed by \specializer effective and efficient
in comparison to the baseline model? (\autoref{sec:eval::specialization})

\item  
How do the model selection policies employed by \selector cope with drift?
(\autoref{sec:eval::selection})

\item 
\edit{
How does \sys perform in the presence of drift? (\autoref{sec:eval::full}) 
}

\item 
\edit{
How does \sys execute end-to-end queries?
(\autoref{sec:eval:queries})}

\item 
\edit{
What is the impact of each component of \sys on its efficacy?
(\autoref{sec:eval:ablation}) }

\squishend

\subsection{System Setup}
\label{sec:eval::setup}

\PP{Implementation}
\label{sec:eval::setup:::implementation}
We implement \sys in Python 3.6.
We develop all of the convolutional neural networks using PyTorch
1.4~\cite{pytorch}.
We leverage an off-the-shelf implementation of YOLOv3, and modify its 
layers to construct YOLO-Tiny. 
We use the MS-COCO API to operate on the \bdd dataset~\cite{cocoapi}.

\PP{Machine}
\label{sec:eval::setup:::machine}
We perform our experiments on a server with an NVIDIA Tesla P100 (16~GB RAM)
and an Intel Xeon 2GHz CPU (2 threads). 
The server contains 12~GB of RAM.

\PP{Datasets}
\label{sec:eval::setup:::datasets}
We use the following datasets to evaluate \sys.

\noindent
\dcircle{1} \textbf{\mnist:}
\label{sec:mnistdata}
This dataset consists of 60K $28\times 28$ black-and-white images of
handwritten digits\cite{mnist}.
We use this dataset to highlight the properties of the latent space associated
with standard and adversarial AEs
in~\autoref{fig:aelatent} and ~\autoref{fig:aaelatent}.
We validate the drift detection algorithm on this dataset
in~\autoref{sec:eval::detection}.

\noindent
\dcircle{2} \textbf{\cifar:}
\label{sec:cifar10data}
This dataset consists of 60K $32\times 32$ colored images belonging to ten
classes\cite{cifar}. 
We also use this dataset to validate the drift detection algorithm.

\noindent
\dcircle{3} \textbf{\bdd:}
\label{sec:bdddata}
This dataset consists of 100K $1280 \times 720$ colored images obtained from
dashboard camera videos~\cite{bdd}.
These high-resolution images are captured under diverse environments:

\squishitemize
\item \textbf{Time of day:} dawn, day, and night.
\item \textbf{Weather:} rainy, snowy, foggy, cloudy, and overcast conditions.
\item \textbf{Location:} residential, highway, city, and other
locations.
\squishend

There are ten classes of objects in \bdd (\eg, traffic lights, cars).
%
%
%
%
%
%
\edit{
We use this dataset to validate the efficacy of \sys across drifting
environmental conditions.
We initially train models on a subset of the \bdd dataset.
We then introduce unseen clusters in \bdd during our evaluation to examine
\sys{'s} drift detection and recovery capabilities.
When \detector detects drift due to the novelty of unseen clusters, 
the \specializer starts generating new models for these subsets and the
\selector picks them once they are generated.
} 

\PP{Dimensionality} 
Each dataset's dimensionality is the number of pixels in
each image.
The dimensionality of images in \mnist, \cifar, and \bdd is 784, 1024,
$\sim$921K, respectively.

\subsection{Drift Detection}
\label{sec:eval::detection}

\begin{table}[t]
\small
\caption{
\textbf{Impact of Distance Metric on Drift Detection Accuracy:}
We compare the accuracy of DA-GAN (DG) in \detector against approaches on \mnist and \cifar: 
(1) LOF, (2) DRAE,
(3) AE, (4) adversarial AE (AAE), and PCA.
We reproduce the accuracy scores of LOF\protect\cite{lof} and
DRAE\protect\cite{drae}.
}
\label{tab:mnistperf}
\scalebox{0.9}{
\begin{tabular}{|l|rrrrrr|rrr|}
\hline
\multicolumn{1}{|c|}{\multirow{2}{*}{Outliers}} & \multicolumn{6}{c|}{\textbf{\mnist}}        
& \multicolumn{3}{c|}{\textbf{\cifar}}                                                 \\
\multicolumn{1}{|c|}{}                             & \multicolumn{1}{c}{LOF} & \multicolumn{1}{c}{DRAE} & \multicolumn{1}{c}{AE} & \multicolumn{1}{c}{AAE} & \multicolumn{1}{c}{PCA}  & \multicolumn{1}{c|}{DG} & \multicolumn{1}{c}{AE} & \multicolumn{1}{c}{AAE} & \multicolumn{1}{c|}{DG} \\ \hline
0\%                                               & 0.95                   & 0.98                    & 0.98                     & 0.98                    & 0.82                       & 0.99                       & 0.91                  & 0.98                    & 0.99                      \\
10\%                                               & 0.92                  & 0.95                    & 0.93                     & 0.97                    & 0.69 & 0.98                       & 0.84                  & 0.97                    & 0.97                      \\
20\%                                               & 0.83                  & 0.91                    & 0.90                     & 0.83                    & 0.61 & 0.97                       & 0.82                  & 0.95                   & 0.97                      \\
30\%                                               & 0.72                  & 0.88                    & 0.87                     & 0.91                    & 0.31  & 0.96                       & 0.81                  & 0.93                   & 0.95                      \\
40\%                                               & 0.65                  & 0.82                    & 0.84                     & 0.91                    & 0.30 & 0.95                       & 0.77                  & 0.91                   & 0.95                      \\
50\%                                               & 0.55                  & 0.73                    & 0.82                     & 0.90                    & 0.29 & 0.94                       & 0.74                  & 0.89                   & 0.94                      \\ \hline
\end{tabular}
}
\end{table}

In this experiment, we measure the efficacy of \detector
(\autoref{sec:driftdetection}).
We first compare the F1-score of the DA-GAN distance metric on two datasets
against that delivered using other distance metrics.
We compare DA-GAN against two state-of-the-art distance metrics that are geared
towards low-dimensional data:
(1) LOF~\cite{lof}, and (2) DRAE~\cite{drae}. 
\edit{We also compare it against PCA, a canonical dimensionality reduction
technique~\cite{zou2006sparse}.}
LOF estimates density of the input space and clusters regions of similar
density.
It detects drift by comparing the density distribution of recent points to that
of the training data.
DRAE uses the reconstruction error of an AE on the high-dimensional
output images to detect drift.

Since DA-GAN models the distribution using a low-dimensional manifold for
identifying drift, it is more robust to the curse of dimensionality.
We configure $\Delta=0.75$ in DA-GAN (\autoref{fig:deltaband}).
We train DA-GAN for 100 epochs with a learning rate of 0.003 using the
adversarial training procedure in~\autoref{alg:training}.

\begin{table*}[t]
\small
\caption{
\textbf{Distribution of Images:}
We compute the distribution of the images in the \bdd dataset across the four 
clusters identified by \detector in an unsupervised manner based on their true
labels. A subset of images in the dataset are not labeled (\textit{undefined}).}
\label{tab:demographics}
\begin{tabular}{|l|rrr|rrr|rrr|rrr|rrr|c|}
\hline
\multicolumn{1}{|c|}{\multirow{2}{*}{Clusters}} &
\multicolumn{3}{c|}{\textbf{Clear}}                               
& \multicolumn{3}{c|}{\textbf{Foggy}}                              
& \multicolumn{3}{c|}{\textbf{Overcast}}                                                   
& \multicolumn{3}{c|}{\textbf{Rainy}}                                                     
& \multicolumn{3}{c|}{\textbf{Snowy}}                                                       
& \multicolumn{1}{c|}{\textbf{Undefined}} 
\\
\multicolumn{1}{|c|}{} &
\multicolumn{3}{c|}{(57428 imgs)}                               
& \multicolumn{3}{c|}{(143 imgs)}                              
& \multicolumn{3}{c|}{(10009 imgs)}                                                   
& \multicolumn{3}{c|}{(5795 imgs)}                                                     
& \multicolumn{3}{c|}{(6316 imgs)}                                                       
& \multicolumn{1}{c|}{(20309 imgs)} 
\\
\multicolumn{1}{|c|}{}                          & \multicolumn{1}{c}{Dawn} & \multicolumn{1}{c}{Day} & \multicolumn{1}{c|}{Night} & \multicolumn{1}{c}{Dawn} & \multicolumn{1}{c}{Day} & \multicolumn{1}{c|}{Night} & \multicolumn{1}{c}{Dawn} & \multicolumn{1}{c}{Day} & \multicolumn{1}{c|}{Night} & \multicolumn{1}{c}{Dawn} & \multicolumn{1}{c}{Day} & \multicolumn{1}{c|}{Night} & \multicolumn{1}{c}{Dawn} & \multicolumn{1}{c}{Day} & \multicolumn{1}{c|}{Night} & \multicolumn{1}{c|}{}                           \\ \hline
\textbf{\cmday}                                      & 90\%                    
& 99\% & 0\%                        & 0\%                      & 21\%                    & 0\%                        & 58\%                     & 75\%                    & 0\%                        & 0\%                      & 10\%                    & 0\%                        & 15\%                     & 1\%                     & 0\%                        & 61\%                                           \\
\textbf{\cmnight}                                      & 0\%                     
& 0\%                     & 100\%                      & 0\%                      & 0\%                     & 100\%                      & 0\%                      & 0\%                     & 100\%                      & 0\%                      & 6\%                     & 100\%                      & 0\%                      & 0\%                     & 100\%                      & 35\%                                           \\
\textbf{\cmrainy}                                      & 0\%                     
& 0\%                     & 0\%                        & 63\%                     & 23\%                    & 0\%                        & 41\%                     & 18\%                    & 0\%                        & 100\%                    & 80\%                    & 0\%                        & 28\%                     & 0\%                     & 0\%                        & 3\%                                            \\
\textbf{\cmsnowy}                                      & 9\%                     
& 1\%                     & 0\%                        & 38\%                     & 57\%                    & 0\%                        & 1\%                      & 7\%                     & 0\%                        & 0\%                      & 3\%                     & 0\%                        & 57\%                     & 99\%                    & 0\%                        & 0\%                                            \\ \hline
\end{tabular}
\end{table*}

\PP{MNIST} 
We configure two digits to be outlier classes.
We vary the percentage of outliers in the test dataset
from 0\% through 50\%.
The results are shown in~\autoref{tab:mnistperf}.
\edit{
The most notable observation is that LOF and DRAE metrics do not scale up even
to the comparatively low-dimensional images in \mnist. 
PCA exhibits lower accuracy since it does not take the spatial locality of
the pixels in the image into consideration.}
As we increase the percentage of outliers to 50\%, the accuracy of LOF and DRAE
drops to 0.73 and 0.55, respectively, and PCA drops to 0.28\%.
In case of LOF, we attribute this to its reliance on the nearest neighbor
distance.
In case of DRAE, this is because it directly uses the reconstruction error on
the output images.
\edit{DA-GAN detects outliers more effectively by projecting the inputs to a
low-dimensional manifold, since it captures the information in the image with
the GAN.}
As we increase the percentage of outliers to 50\%, accuracy of DA-GAN only
drops from 0.99 to 0.94.

\begin{figure}[t] 
    \centering \includegraphics[width=\textwidth]{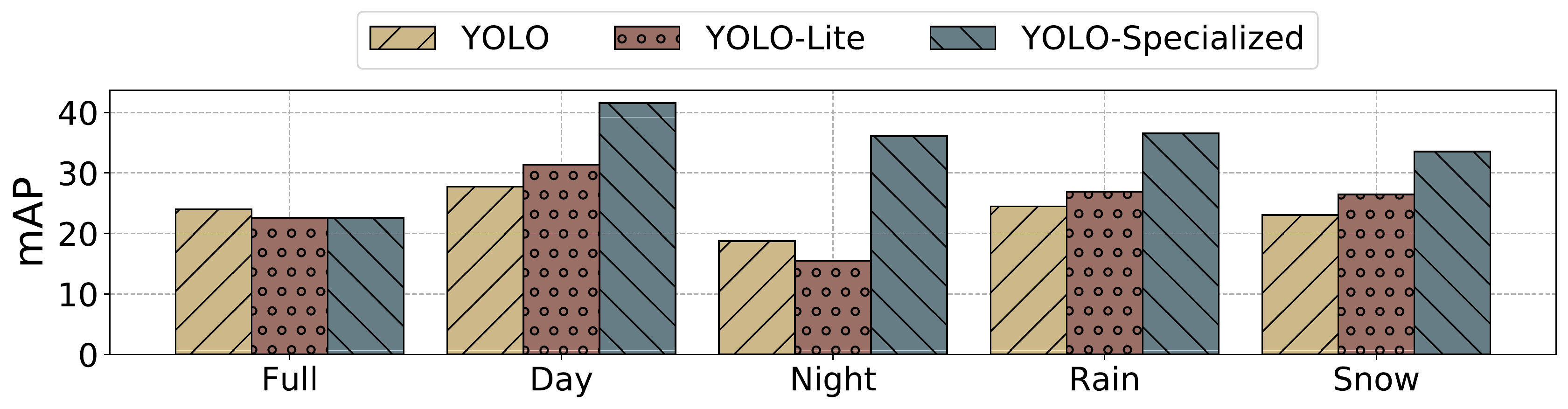}
	\caption{\textbf{Impact of Model Specialization on Accuracy}: 
	We compare the detection accuracy (mAP metric) of the static \yolo model against
	the models constructed by \specializer: \yololite and \yololocal.
	}
	\label{fig:locallite}
\end{figure}

\PP{CIFAR-10} 
We conduct a similar empirical analysis on \cifar.
Prior work on outlier detection has only focused on cross-class accuracy
differences in this dataset (and not on the percentage of outliers).
So, we compare DA-GAN against AE and adversarial AE (AAE) distance metrics.
These metrics outperform LOF and DRAE metrics on \mnist.
For instance, on \mnist, when the percentage of outliers is 50\%, the accuracy
of AE and AAE metrics only drop to 0.82 and 0.90, respectively.
On \cifar, the accuracy of AE and AAE metrics drop to 0.74 and 0.89,
respectively.
We attribute this to the higher dimensionality of images in \cifar compared to
that in \mnist.
The adversarial AE metric outperforms its standard counterpart by circumventing 
the irregular mapping problem (\autoref{sec:preliminaries::generative}).
DA-GAN outperforms AE and AAE metrics on this dataset.
As we increase the percentage of outliers to 50\%, its accuracy only
drops from 0.99 to 0.94.
%
%

\PP{BDD} 
We next evaluate the efficacy of \detector on the \bdd dataset.
This is a challenging dataset with a high-dimensional manifold.
We only use the DA-GAN metric in this experiment.
We train the DA-GAN on a held-out subset of \bdd consisting of $\sim$20~K
images.
These images do not have any time of day or weather labels associated with 
them (\ie, \ttt{undefined} images in~\autoref{tab:demographics}).

We seek to examine its ability to detect images from previously unseen classes.
The dataset contains 15 labeled subsets based on different environmental
conditions.
We note that the time of day and weather attributes of a video are independent
(\eg, video collected on a snowy night)\footnote{\detector found that the
location attribute is not important from a drift detection standpoint.}.
We construct a workload that exhibits gradual drift by introducing images from
the outlier subsets.

\detector identifies drift using unsupervised clustering with the DA-GAN
distance metric (\ie, without using the time of day and weather attributes of
images).
%
%
It automatically learns four clusters out of the 15 subsets. 
The results are summarized in~\autoref{tab:demographics}.
We compute the distribution of the images across the detected clusters based on
their labels to examine why \detector picked these clusters.
\cday mostly contains images captured on clear days as well as a few overcast
images that are tagged as partially cloudy.
\detector groups nearly all of night-time images into \cnight. 
\crainy mostly contains images with rain (as well as some images with snowfall
and fog).
\csnowy mostly contains images with snowfall along with a few images with fog
(\eg, \textit{fog-day}, \textit{fog-night} pairings).
The distribution of images across these clusters indicate that
\detector identifies the key features of the dataset.
Among the 15 labeled subsets of the \bdd dataset, 
\detector automatically subsumes similar subsets into the same cluster.
For instance, it maps nearly all of the night-time images ($\sim$ 98\%) to
\cnight, irrespective of the weather condition.

\begin{table}[t]
\small
\caption{\textbf{Impact of Model Specialization on Cross-Subset Detection Accuracy:}
We compare the cross-subset detection accuracy of the \yolo model against the
models constructed by \specializer: \yololocal and \yololite.
}
\begin{tabular}{|l|lllll|}
\hline
\multicolumn{1}{|c|}{\multirow{2}{*}{Data}} & \multicolumn{5}{c|}{\textbf{Cluster used for Specialization}}                                                  \\
\multicolumn{1}{|c|}{}                      & Baseline   & \cmday             & \cmnight           & \cmrainy           & \cmsnowy           \\ \hline
\fulldata                                        & {\ul 0.2403} & 0.2068          & 0.2215          & \textbf{0.2581} & 0.2445          \\ \hline
\daydata                                        & 0.2772       & \textbf{0.4157} & 0.2229          & 0.2900          & 0.3339          \\
\nightdata                                      & 0.1875       & 0.0789          & \textbf{0.3609} & 0.2691          & 0.2439          \\
\rainydata                                      & 0.2449       & 0.2424          & 0.2645          & \textbf{0.3656} & 0.3223          \\
\snowydata                                      & 0.2304       & 0.2082          & 0.2467          & 0.2636          & \textbf{0.3354} \\ \hline
\end{tabular}

\label{tab:performances}
\end{table}

\PP{BDD Clusters}
Using the clusters obtained in this experiment, we construct five data subsets
that we leverage for testing in later experiments:
(1) all of the images (\fulldata, 79863 images
\footnote{\bdd contains three splits of 69863, 20137, and 10000 images each. We
set aside the second split to train non-specialized models.}), 
(2) images captured during day-time under clear weather conditions (\daydata ,
40696 images), 
(3) images captured during night-time under any weather condition (\nightdata,
31900 images), 
(4) images captured under rainy or overcast weather conditions (\rainydata, 5808
images), and
(5) images captured under snowy weather conditions (\snowydata, 6313 images).
%

%
%
%
%
%

\subsection{Model Specialization}
\label{sec:eval::specialization}

In this experiment, we examine the efficacy of the models constructed by the
\specializer in \sys for each of the four detected
clusters.
%

\PP{Specialized vs. Lite models} 
We first examine the detection  accuracy of the three object detector models
(\autoref{sec:yolo}): \yolo, \yololocal, and \yololite.
We train and test the models over the five \bdd clusters.
%

\PP{Detection Accuracy}
The results are shown in~\autoref{fig:locallite}.
The most notable observation is that \yololocal is the best performing model
across all subsets (except for \fulldata).
For each cluster, \yololocal delivers higher detection accuracy that its counterparts
since it is specialized only on that subset. 
For example, on \nightdata, the \yololocal model delivers 2$\times$ higher
accuracy compared to \yolo.
It improves accuracy by 1.5$\times$ on average across all clusters.

\specializer directly trains the \yololite student model using the outputs of
\yolo without requiring externally sourced labels.
So its detection accuracy is comparable to \yolo across most subsets.
\yololite{'s} detection accuracy on \nightdata than that of \yolo.
This is because \yolo makes most of its mistakes on this challenging subset. 
Since \yololite is smaller than \yolo, it does not learn all of the features 
of \nightdata.

\begin{table}[t]
\small
\caption{\textbf{Impact of Model Specialization on Performance and Memory
Footprint}:
We compare the performance and memory footprint of the baseline \yolo model
against the models constructed by \specializer: \yololocal and \yololite.
}
\label{tab:yolostats}
\begin{tabular}{|llll|}
\hline
\textbf{Model}          & \textbf{Architecture}\cite{yolo}                                                  
& \textbf{Throughput} & \textbf{Size}  \\ \hline \yolo           &
YOLOv3 & 24 FPS     & 237MB \\
\yololocal & \begin{tabular}[c]{@{}l@{}}Pruned YOLOv3-tiny\end{tabular} & 144 FPS    & 34MB  \\
\yololite      & YOLOv3-tiny                                                   
& 140 FPS    & 35MB  \\ \hline
\end{tabular}
\end{table}

\PP{Cross-Subset Detection Accuracy}
We next examine the detection accuracy (\ie, mAP score) of the specialized \yololocal
models on other subsets that they are not trained on.
Due to class imbalance in the \bdd dataset, we train each model on the
same number of samples (constrained by the smallest cluster).
As shown in~\autoref{tab:performances}, each specialized model outperforms the
 \yolo model on their target subset.
For instance, consider the \yololocal model trained on \cday.
On \daydata, it delivers 2$\times$ higher detection accuracy than the model tailored
for \cnight.
It also works well on \rainydata and \snowydata since most of the data in
these subsets are taken during the day.
However, it delivers 5$\times$ lower detection accuracy on the \nightdata in comparison 
to \cnight.
This is because most of the training data for \cday are captured on clear
days, which is different from
\nightdata.

\PP{Model Generation Time}
\edit
{
Since \sys automatically clusters the dataset, each cluster contains more
homogenous data points compared to the entire dataset.
So, it is able to quickly generate smaller \yololite and \yololocal models on
these clusters.
For example, on \nightdata, \sys generates a \yololocal model 21$\times$ faster
compared to an off-the-shelf unspecialized \yolo model.
The reasons for this are twofold.
First, the specialized model consists of 7$\times$ fewer parameters compared to
the original \yolo model.
Second, the \nightdata cluster contains 3$\times$ fewer images compared to
\fulldata.
Thus, reduction in model and dataset sizes enable faster training of
specialized models.}

\PP{Query Execution Time}
\edit{
As shown in~\autoref{tab:yolostats}, \sys delivers higher throughput by
leveraging these models.
Since  \yololocal and \yololite are 7$\times$ smaller than \yolo, they are
nearly 6$\times$ faster than the \yolo model.
}

\PP{Memory Footprint}
Since \specializer constructs four models on the \bdd dataset, 
the overall memory footprint of \sys is 2$\times$ smaller than with the baseline
\yolo model.

When \detector finds a new cluster, \specializer first generates a \yololite
model using the outputs of \yolo.
\yololite{'s} advantage over \yololocal lies in faster training as there is no
need to wait for externally sourced labels.
When the labels are available, either from humans or using 
weak-supervision ~\cite{snorkel}, 
\specializer constructs a \yololocal model that delivers higher detection accuracy than
its \yololite counterpart.
In the rest of the experiments, we configure \sys to use \yololocal models.

\begin{figure*}[t] 
    \centering \includegraphics[width=\textwidth]{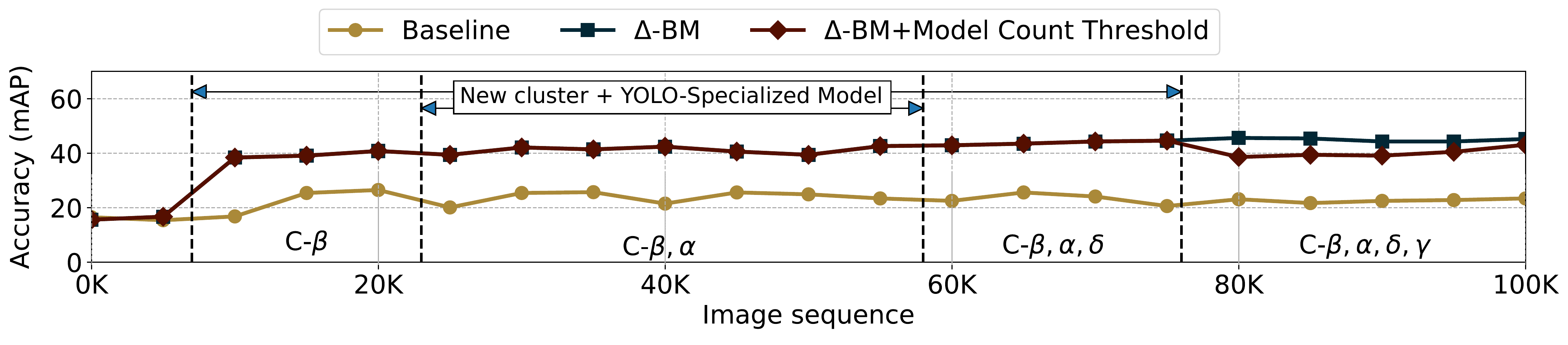}
	\caption{\footnotesize{\textbf{End-to-End Evaluation}: 
	We examine detection accuracy of \sys with all components under three configurations.
	\dcircle{1} \textbf{Baseline}
	A large \yolo model is used to process all \bdd videos.
	\dcircle{2} \textbf{$\Delta$-BM:} 
	We enable drift recovery and configure \sys to use the $\Delta$-BM selection
	policy.
	\dcircle{3} \textbf{$\Delta$-BM + Model Count Threshold:}
	We next limit the maximum number of models to three.
	}}
	\label{fig:runtimes}
\end{figure*}

\subsection{Model Selection}
\label{sec:eval::selection}

In this experiment, we compare the efficacy of the model selection policies 
discussed in~\autoref{sec:selector} on the \bdd dataset: 

\dcircle{1} 
\textbf{KNN-U:} \selector uses the unweighted average of the four models
constructed by \specializer.

\dcircle{2} 
\textbf{KNN-W:} \selector uses the weighted average of the four models
constructed by \specializer.
\selector computes weights by normalizing the distances 
in the latent space obtained using DA-GAN using ~\autoref{eq:weighting}.
%

\dcircle{3} 
\textbf{$\Delta$-BM:} 
With this policy, \selector uses the high-density $\Delta$-bands of cluster while picking models.
%
For an image that falls inside a $\Delta$-band, we select the model associated
with that $\Delta$-band's cluste.
%
For an image that falls outside any of existing $\Delta$-bands, we revert
to the KNN-W policy (8\% of the images in \bdd).
For an image the falls inside multiple overlapping $\Delta$-bands, we use all
of the bands with equal weights (39\% of the images in \bdd).

\begin{table}[t]
\small
\caption{\textbf{Impact of Model Selection on Accuracy:}
We compare the policies adopted by \selector for picking the specialized
\yololocal models compared to the baseline \yolo model.
}
\begin{tabular}{|l|llll|}
\hline
\multicolumn{1}{|c|}{\multirow{2}{*}{Data}} & \multicolumn{4}{c|}{\textbf{Model Selection Policy}}                                                  \\
\multicolumn{1}{|c|}{}                      & Baseline   &  KNN-U           &
KNN-W & $\Delta$-BM           \\ \hline \fulldata                                        & 0.2403      & 0.2365          & \textbf{0.2811} & 0.2491          \\ \hline
\daydata                                        & 0.2772       & 0.3514          & 0.3954          & \textbf{0.4257}          \\
\nightdata                                      & 0.1875       & 0.2123          & 0.3432          & \textbf{0.3687}          \\
\rainydata                                      & 0.2449       & 0.2843          & \textbf{0.3764} & 0.3552          \\
\snowydata                                      & 0.2304       & 0.3134          & 0.3412          & \textbf{0.3653} \\ \hline
\end{tabular}
\label{tab:msp}
\end{table}

The results are shown in ~\autoref{tab:msp}.
Our baseline is the a static system without drift detection or recovery.
KNN-W outperforms KNN-U on all of the subsets.
For instance, on \rainydata, it delivers 32\% higher detection accuracy compared to
KNN-U.
This is because it ensures that the best-fit model (\ie, the one specialized on
\crainy) is given the highest consideration.

$\Delta$-BM policy outperforms KNN-W on most of the subsets.
For instance, on \daydata, it delivers 7.5\% higher detection  accuracy compared to
KNN-W.
We attribute this to $\Delta$-BM policy's focus on high-density bands instead of the entire
clusters (as KNN-U and KNN-W do).
This policy works well in tandem with \detector that leverages $\Delta$-bands to
identify drift.
%
%
On \rainydata, KNN-W outperforms $\Delta$-BM by 6\%.
This is because $\Delta$-BM only uses the high-density bands of \crainy,
since it contains most images with rain.
However, \crainy does not contain images with cloudy skies.
So the model trained on this cluster is slightly less effective on \rainydata .
KNN-W circumvents this limitation by using all of the modelss.

\subsection{End-to-End Evaluation}
\label{sec:eval::full}

We next examine the efficacy and efficiency of all of the components of \sys 
in tandem.
We evaluate \sys under three configurations on a sequence of 100~K images in
\bdd. 
We construct the sequence thus:
(1) 20~K images exclusively from \nightdata images,
(2) after 20~K images, we add \daydata to the pool. 
(3) after 40~K images, we add \snowydata to the pool, and
(4) after 60~K images, we add \rainydata to the pool.
The chance for selecting an image of any subset is not adjusted for equal chance,
since we want to replicate a realistic distribution.
We measure the object detection accuracy (mAP) of \sys every
5~K images in the sequence.
The results are shown in~\autoref{fig:runtimes}.  

\dcircle{1} \textbf{Baseline:} 
In the baseline configuration, \sys uses a single \yolo model to process the
entire sequence of images.
Without drift recovery, this system's detection accuracy is
$\sim$20 mAP.
This is because it is unable to detect and recover from drift.
\sys processes images at 24~FPS under this configuration.
Since there are no specialized lightweight models, its performance is
constrained by the throughput of the heavyweight \yolo model (see ~\autoref{tab:yolostats})

\dcircle{2} \textbf{$\Delta$-BM:} 
We next enable drift recovery and configure \sys to use the $\Delta$-BM
selection policy.
In this configuration, \sys first uses a \yololite model to process the
\nightdata images. 
The accuracy is comparable to the baseline.
This is because \yololite delivers similar accuracy to the full model
(~\autoref{fig:locallite}).   
When \detector identifies a new cluster, \sys generates a \yololocal 
model and switches to it (as it outperforms the \yololite and full models).
%
Each of the dotted lines in~\autoref{fig:runtimes} represents identification 
of a new cluster by \detector and  subsequent generation of a \yololocal model.
The specialized models double the detection accuracy from $\sim$20 mAP to
$\sim$40 mAP.
This is because the \selector picks the appropriate model
constructed by \specializer using the $\Delta$-BM policy.
%
%
%
%
%
%

\dcircle{3} \textbf{$\Delta$-BM + Model Count Threshold:}
We next limit to the maximum number of models to three.
When \detector identifies the fourth cluster (\ie, \cmrainy), it drops the
cluster with the smallest number of inputs.
In this dataset, it drops \csnowy (5~K images in cluster).
Since it only relies on the three other models for prediction, its detection accuracy
suffers slightly due to the missing model.
With the $\Delta$-BM policy, \selector reverts to KNN-W when encountering 
points outside the existing $\Delta$ bands.
So, the drop in detection accuracy is not significant.
The throughput is slightly higher due to fewer models (4 $\rightarrow$ 3), at 140~FPS

\subsection{Aggregation Query}
\label{sec:eval:queries}

\begin{table}[t]
\small
\caption{\edit{\textbf{Aggregation queries and Lightweight Filters:} 
We compare the efficacy and efficiency of executing aggregation queries across
several configurations.
These configurations include: 
(1) a static system without specialized models, 
(2) \sysheavy that uses specialized \yolo models,
(3) \sys with no filters and \yololocal models,
(4) \syspp with unspecialized filters and \yololocal models, and
(5) \sysfilter with specialized filters and \yololocal models.
}}
\label{tab:aggregation}
\begin{tabular}{|c|l|l||l|l|l|}
\hline
\multicolumn{1}{|l|}{}                                                           & \textbf{Architecture}       & \textbf{Metric} & \textbf{Cars} & \textbf{Trucks} & \textbf{FPS}         \\ \hline
\multirow{3}{*}{\begin{tabular}[c]{@{}c@{}}Aggregation \\ Queries\end{tabular}}  & Static                      & Query Acc.        & 0.65          & 0.86            & 24                   \\ \cline{2-6} 
                                                                                 & \sys                        & Query Acc.        & 0.94          & 0.92            & 140                  \\ \cline{2-6} 
                                                                                 & \sysheavy                   & Query Acc.        & 0.97          & 0.98            & 20                   \\ \hline
\multirow{4}{*}{\begin{tabular}[c]{@{}c@{}} Aggregation \\ Queries \\ with
Filters\end{tabular}} & \multirow{2}{*}{\sysfilter} & Query Acc.        & 0.92          & 0.83            & \multirow{2}{*}{130} \\
                                                                                 &                             & Reduction  & 8\%           & 68\%            &                      \\ \cline{2-6} 
                                                                                 & \multirow{2}{*}{\syspp}     & Query Acc.        & 0.59          & 0.76            & \multirow{2}{*}{135} \\
                                                                                 &                             & Reduction  & 38\%          & 76\%            &                      \\ \hline
\end{tabular}
\end{table}

\edit{We next examine how \sys complements the filtering technique used in
state-of-the-art visual DBMSs~\cite{blazeit,pp}.
We focus on aggregation queries (\eg, number of cars in a set of videos):}
\begin{lstlisting}[style=SQLStyle] SELECT COUNT(detections)
SELECT FROM (SELECT detections
        FROM bdd USING MODEL yolo_specialized 
        WHERE class='car')
\end{lstlisting}

\edit{We consider two classes: cars and trucks. 
In each case, \selector selects the appropriate  \yololocal model for each
image. 
We compare \sys against two systems:
(1) a static system without specialized models, and
(2) a variant of \sys that uses specialized \yolo models instead of specialized
\yololocal models.
We refer to the latter variant as \sysheavy.
The specialized \yolo models used by \sysheavy are 6$\times$ larger and slower
than \yololocal models.}

\edit{As shown in~\autoref{tab:aggregation}, \sys returns more accurate results 
compared to the static system. 
For cars, \sys and \sysheavy are 50\% better than a static system.
For trucks, which are larger objects, all systems perform better.
%
%
While \sysheavy is slightly more accurate than \sys ($\sim$3-6\%), 
it is 7$\times$ slower.}

\begin{figure}[t] 
	\begin{subfigure}{.75\textwidth}
		\centering
		\includegraphics[width=\linewidth]{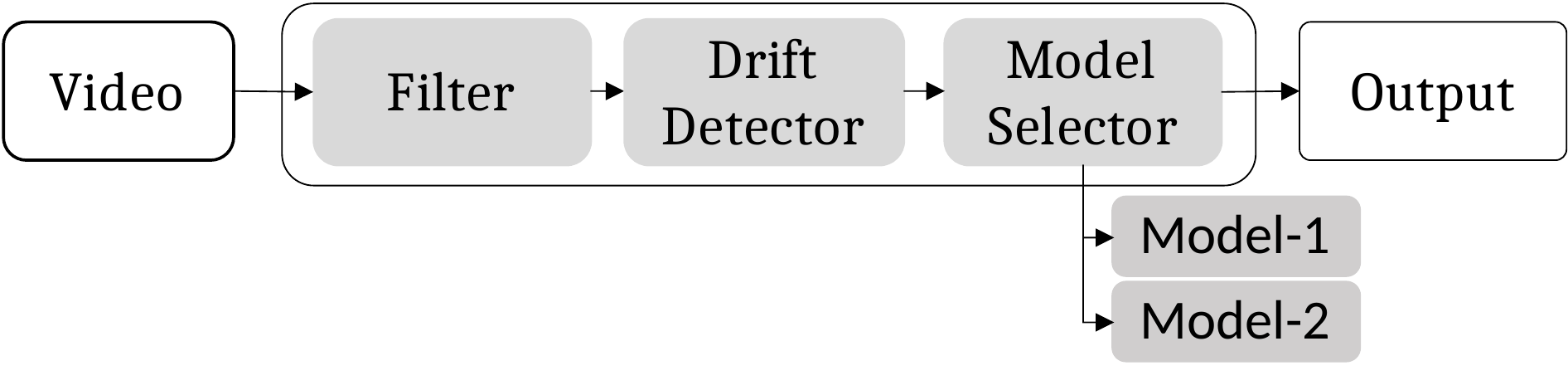}
		\caption{\syspp}
		\label{fig:odinppbasicresponse}
	\end{subfigure}
	\\
	\begin{subfigure}{.75\textwidth}
		\centering
		\includegraphics[width=\linewidth]{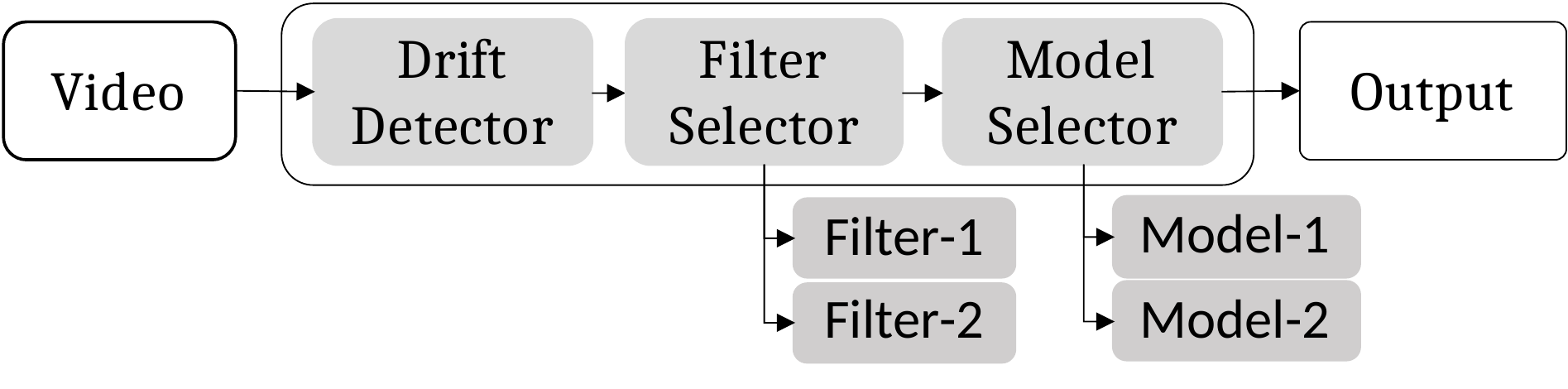}
		\caption{\sysfilter}
		\label{fig:odinppbasicresponse}
	\end{subfigure}
	\\
	\caption{\edit{\textbf{\sys Configurations}: We augment \sys with lightweight 
	DNN filters to improve throughput~\protect\cite{pp}.
    (1) \syspp with specialized models and unspecialized filter.
    (2) \sysfilter with specialized models and specialized filters.
	}}
	\label{fig:odinpp}
\end{figure}

\PP{Aggregation Queries Using Lightweight Filters}
\edit{We next examine how to accelerate the aggregation queries using lightweight
filters~\cite{pp}.
In this case, the system returns approximate aggregates.
\sys creates specialized filters for each cluster. 
We modify the architecture of \sys to incorporate these filters, as shown
in~\autoref{fig:odinpp}.
The filter is a lightweight DNN that preprocesses the images to return a boolean
decision that indicates whether that image must be subsequently processed by the
heavweight model (\eg, \yololocal).
In our example, a DNN with 3 convolutional layers is sufficient to determine 
if a given frame contains a car or not.
If the frame has no cars, then \sysfilter does not process it with the
\yololocal model that counts the number of cars.
In this case, the query looks thus:} 

\begin{lstlisting}[style=SQLStyle]
SELECT COUNT(detections) 
   FROM (SELECT detections 
         FROM (SELECT * FROM bdd 
               USING FILTER car_filter 
               WHERE class=1))
         USING MODEL yolo_specialized 
         WHERE class='car'          
\end{lstlisting}

\edit{We consider three configurations: 
(1) \sys with specialized models and no filters, 
(2) \syspp with specialized models and unspecialized filter~\cite{pp},  and 
(3) \sysfilter with specialized models and specialized filters. 
The results are shown in~\autoref{tab:aggregation}.
With \sysfilter, there is 8\% data reduction for `cars' (since cars are present
in nearly every frame).
Query accuracy slightly drops since the filter returns some false negatives.
With trucks, we observer higher data reduction since they are rarer in \bdd.
The drop in query accuracy is more prominent with \syspp since it uses a single
unspecialized filter.
In the presence of drift, this filter returns more false negatives.
With trucks, the filters miss more frames in \nightdata due to 
lighting conditions.
This experiment shows that drift detection and recovery is important for filters
as well.}
%



\subsection{Ablation Study}
\label{sec:eval:ablation}

\edit{
We next conduct an ablation study to delineate the impact of each component
of \sys.
Since the \detector is not useful without the recovery components, we consider
these configurations:
\squishitemize
\item End-to-End System: With all three components.
\item - \selector: With only the \detector and \specializer components.
\sys uses the most recently created \yololocal model in this setting.
%
%
%
\item Baseline: Lastly, we remove all the three components.
In this configuration, \sys uses the heavyweight \yolo model.
\squishend
}

\edit{
We summarize the results in~\autoref{tab:ablation}. 
Eliminating the \selector leads to a drop in accuracy, since the best model is
no longer used for each cluster. 
The naive model selection policy is only useful when the drift is monotonically
increasing.
In practice, older clusters co-exist with newer clusters, as is the case in
\bdd.
Since the most recent model is trained on newer clusters, its accuracy drops
when older clusters are re-introduced.
%
%
The memory footprint and throughput are nearly unchanged, since the \selector is
computationally lightweight.
Lastly, removing the \detector and \specializer is equivalent to using a static
heavyweight \yolo model.
The lack of specialization leads to lower accuracy.
Furthermore, performance also suffers since the \yolo model is larger and slower
than the \yololocal models constructed by the \specializer.
}

\begin{table}[t]
\small
\caption{\edit{\textbf{Ablation study for \sys}: 
We delineate the impact of each component of \sys.
}}
\label{tab:ablation}
\begin{tabular}{|l|l|l|l|l|}
\hline
Experiment       & mAP & Query Acc & Throughput & Memory \\ \hline
End-to-End Model & 40.15         & 93.5      & 140FPS     & 148MB            \\
\hspace{1pt}-\selector       & 24.84         & 71.4      & 140FPS     & 148MB            \\
\hspace{1pt}Baseline    & 24.03         & 64.6      & 24FPS      & 237MB            \\
\hline
\end{tabular}
\end{table}

\section{Limitations}
\label{sec:discussionlimitation}

%
%
We now discuss the limitations of \sys and present our ideas for
addressing them in the future.

\PP{Availability of Oracle Labels}
In \sys, we assume that oracle labels are available for images in newly detected
clusters.
In practice, these labels may not be available quickly if they are collected
from humans.
\sys could circumvent this problem by first constructing fast \yololite
models using the outputs of the pretrained \yolo model, thereby bypassing the
label availability constraint.
While these models deliver performance comparable to their \yololocal
counterparts, they suffer from lower accuracy on newly detected clusters.
After the labels are obtained, \sys trains \yololocal models and replaces their 
\yololite counterparts with these newly trained models.
Weak supervision techniques may accelerate the
procurement of oracle labels~\cite{snorkel}.

\PP{DA-GAN Performance} 
The performance of DA-GAN drops over time as the number of clusters increase.
This is because it needs to compare each input against all of the $\Delta$-bands
associated with these clusters.
We believe that locality-sensitive hashing~\cite{lsh} might alleviate this
problem.
Another alternative is to design a more efficient model architecture for the
encoder in DA-GAN, thereby reducing the time taken to encode a point.

\vspace{0.15in}

\section{Related Work}
\label{sec:related}

\PP{Drift Detection}
\cite{drift2} presents a survey of several supervised drift detection mechanisms
Unsupervised methods that detect drift based on the expected data distribution
include model confidence methods~\cite{trustclass,md3b} and clustering
algorithms~\cite{olindda,samknn}.
Outlier detection algorithms detect drift in low-dimensional structured data
(\autoref{fig:driftae}).
DRAE~\cite{drae} uses the reconstruction error of an AE to detect drift.
Since AEs suffer from holes in their latent space, DRAE is only effective 
for static low-dimensional datasets.
LOF~\cite{lof} measures the density of the input space and clusters regions of
similar density.
It detects drift by comparing the density distribution of recent points to that
of the training data.
Researchers have also proposed windowing algorithms to adapt models when the
type of drift is not known~\cite{samknn}.
These algorithms use static windows to track changes in distribution.
Unlike these techniques, \sys generalizes to unstructured data.
The reasons for this are threefold.
First, DA-GAN represents high-dimensional data better than the AE
in~\cite{drae}.
Second, $\Delta$-bands compare high-density regions better than kNN
in~\cite{lof}.
Lastly, it dynamically generates clusters over time instead of using static 
windows employed in~\cite{samknn}.

\PP{Model Specialization}
Recovering from drift is key to maintaining the accuracy of the overall system.
\sys relies on model specialization for drift recovery.
It deploys models specialized for each detected cluster of the data
space.
Model distillation is a widely-used technique for 
specialization~\cite{distillation,teacher}.
With distillation, a teacher model trains a lite (\ie, smaller and faster)
student model to mimic its output. 
It is useful in scenarios where the teacher model is unlikely to fail
(\ie no drift). 
Model compression is another technique for specialization~\cite{compression}.
With compression, we start with a pre-trained model and prune weights below a
threshold to reduce size.
A pre-trained model is not effective in the presence of drift.
Different from these techniques, \sys relies on specialized models for specialization,
where the models are trained from scratch on the novel data points. 
This enables it to work well on drifting datasets.

\PP{Model Selection}
Given a collection of specialized models, it is important to choose the
appropriate ones for processing an input.
Prior efforts on model selection are geared towards low-dimensional data.
ARF constructs an ensemble of weak decision trees and dynamically prune trees
whose accuracy degrades due to drift~\cite{arf}.
It uses a simple majority technique to weight the ensemble of trees.
KME combines several drift detectors to identify cyclical, real, and gradual
drift occurrences~\cite{kme}.
It updates the models if it detects drift or if enough training data is
collected for an update, and assigns weights using a model-to-concept mapping.
It assigns higher weights to models that have been identified to deliver higher
accuracy on recent concepts.
These methods do not work well on high-dimensional data.
\sys uses \selector, which uses either the $\Delta$-DM policy
for picking an ensemble of specialized models for processing a given input.
%

\section{Conclusion}
In this paper, we presented the architecture of \sys, a system for detecting and
recovering from drift in visual data analytics.
We presented an unsupervised algorithm for drift detection by determining the
high-density regions of the input data space.
We proposed the DA-GAN distance metric that allows the \detector to work well
on high-dimensional data.
\sys constructs smaller, faster specialized models for each detected cluster
that deliver higher accuracy compared to the larger, slower model trained on the
entire dataset.
Our evaluation shows that \sys delivers higher throughput, higher detection and query
accuracy, as well as a smaller memory footprint compared to the static setting 
without drift detection and recovery.

\newpage

\newcommand{\newblock}{}
\bibliographystyle{abbrv}
\bibliography{main}

\end{document}